\documentclass[10pt,twocolumn,letterpaper]{article}

\usepackage[pagenumbers]{cvpr} %

\usepackage{graphicx}
\usepackage{amsmath}
\usepackage{amssymb}
\usepackage{booktabs}

\usepackage{dsfont}
\usepackage{mathtools}
\usepackage{bm}

\usepackage{multirow}
\usepackage{colortbl}
\usepackage{amssymb}%
\usepackage{pifont}%
\usepackage{colortbl}
\usepackage[font=small]{subcaption}
\usepackage{enumitem}
\usepackage{hhline}
\usepackage{bbold}
\usepackage{algorithmic}
\usepackage[ruled, linesnumbered]{algorithm2e}
\usepackage[accsupp]{axessibility}
 \usepackage{balance}

\newcommand{\xmark}{\ding{55}}
\newcommand{\cmark}{\ding{51}}
\newcommand{\bestresult}[1]{\textbf{\textcolor{red}{#1}}}
\newcommand{\secondbest}[1]{\textcolor{blue}{\underline{#1}}}

\newcommand{\venueTT}[1]{{$_{\texttt{\text{#1}}}$}}
\newcommand{\venue}[1]{{$_{{\text{#1}}}$}}

\newcommand{\supp}[1]{{\color{blue}  #1}}

\usepackage[pagebackref,breaklinks,colorlinks]{hyperref}

\usepackage[capitalize]{cleveref}
\crefname{section}{Sec.}{Secs.}
\Crefname{section}{Section}{Sections}
\Crefname{table}{Table}{Tables}
\crefname{table}{Tab.}{Tabs.}

\def\cvprPaperID{1154} %
\def\confName{CVPR}
\def\confYear{2023}

\begin{document}

\title{TimeBalance: Temporally-Invariant and Temporally-Distinctive Video Representations for Semi-Supervised Action Recognition}

\author{Ishan Rajendrakumar Dave, Mamshad Nayeem Rizve,  Chen Chen, Mubarak Shah\\
Center for Research in Computer Vision, University of Central Florida, Orlando, USA\\
{\tt\small \{ishandave, nayeemrizve\}@knights.ucf.edu,  \{chen.chen, shah\}@crcv.ucf.edu}}

\maketitle

\begin{abstract}

Semi-Supervised Learning can be more beneficial for the video domain compared to images because of its higher annotation cost and dimensionality. Besides, any video understanding task requires reasoning over both spatial and temporal dimensions. In order to learn both the static and motion related features for the semi-supervised action recognition task, existing methods rely on hard input inductive biases like using two-modalities (RGB and Optical-flow) or two-stream of different playback rates. Instead of utilizing unlabeled videos through diverse input streams, we rely on self-supervised video representations, particularly, we utilize temporally-invariant and temporally-distinctive representations. We observe that these representations complement each other depending on the nature of the action. Based on this observation, we propose a student-teacher semi-supervised learning framework, TimeBalance, where we distill the knowledge from a temporally-invariant and a temporally-distinctive teacher. Depending on the nature of the unlabeled video, we dynamically combine the knowledge of these two teachers based on a novel temporal similarity-based reweighting scheme. Our method achieves state-of-the-art performance on three action recognition benchmarks: UCF101, HMDB51, and Kinetics400. Code: \url{https://github.com/DAVEISHAN/TimeBalance}.

\end{abstract}

\section{Introduction}

Recent development in action recognition have opened up a wide range of real-world applications: visual security systems~\cite{gabv2, rizve2021gabriella,cmu2020}, behavioral studies~\cite{behaviour}, sports analytics~\cite{li2021multisports}, elderly person fall detection systems~\cite{buzzelli2020vision, zhang2012privacy, liu2020privacy}, etc. Most of these developments are mainly courtesy of large-scale curated datasets like Kinetics~\cite{kinetics}, HVU~\cite{hvu}, and HACS~\cite{hacs}. However, labeling such a massive video dataset requires an enormous amount of annotation time and human effort. 
At the same time, there is a vast amount of unlabeled videos available on the internet. The goal of semi-supervised action recognition is to use such large-scale unlabeled dataset to provide additional supervision along with the labeled supervision of the small-scale dataset.

\begin{figure*}[h]
    \centering
    \includegraphics[width=\textwidth]{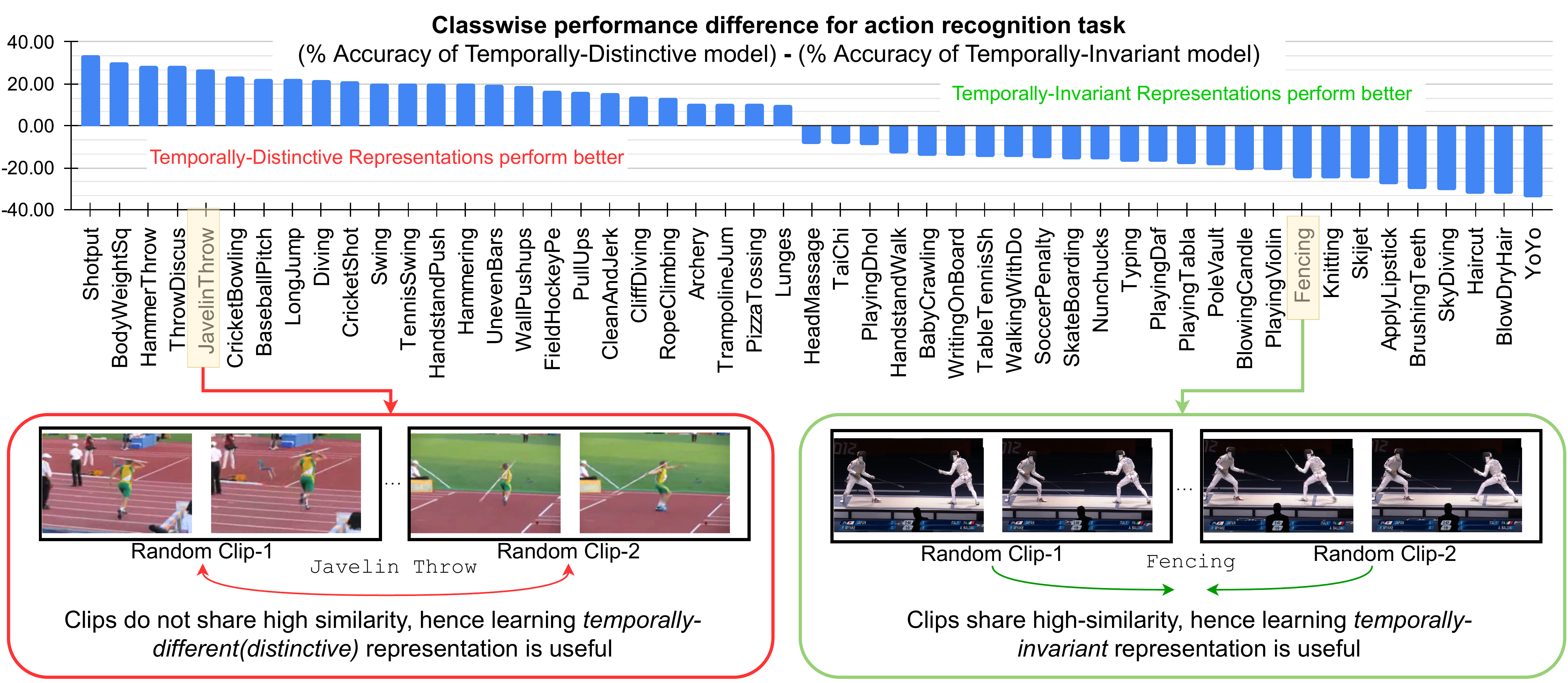}
    \caption{\textbf{Motivation for Temporally-Distinctive and Temporally-Invariant  Representations.} In order to leverage the unlabeled videos effectively, we consider two kinds of self-supervised video representation learning techniques with complementary goals: (1) \textit{Temporally Invariant Representations} (\texttt{Bottom Right}) encourage learning the commonalities of the clips, hence it mainly focuses on learning features related to highly frequent repetitions and appearance. (2) \textit{Temporally Distinctive Representations} (\texttt{Bottom Left}) encourage learning the dissimilarities between clips of the same video, hence it encourages learning features for sub-actions within the video. The plot shows the activity-wise UCF101 performance difference of finetuned models which were self-supervised pretrained with temporally-distinctive and temporally-invariant objectives. The plot shows extreme 25-25 classes after sorting the all classwise differences.}
    \label{fig:motivation}
\end{figure*}

Semi-supervised learning for image classification has seen tremendous progress in recent years~\cite{tarvainen2017mean,Shi_2018_ECCV,arazo2020pseudo,ups}. In semi-supervised action recognition, recent approaches have adapted these image-based methods by incorporating motion-related inductive biases into the setup.
For instance, some methods~\cite{semi_tgfixmatch, semi_mvpl} use two different input modalities where the original RGB video promotes learning appearance-based features while optical flow/temporal gradients promotes learning of motion-centric features. Another set of methods uses input streams of different sampling rates to achieve this~\cite{semi_tcl, semi_tacl}. Although these input-level inductive biases are simple-yet-very-effective to provide unlabeled supervision for action recognition, they are not suitable for large-scale datasets due to their multiplicative storage requirement and high preprocessing overhead.

Contrastive Self-supervised Learning (CSL) has emerged as a powerful technique to learn meaningful representations from unlabeled videos. Existing video CSL methods deal with mainly two different kinds of objectives: (1) Learning similarities across clips of the same video i.e \textit{temporally-invariant} representations~\cite{cvrl, byol, videomoco} (2) Learning differences across clips of the video i.e. \textit{temporally-distinctive} representations~\cite{jenni2021time, tclr, dsm}. 
Each objective has its own advantages, depending on the nature of the unlabeled videos being used. 

Our experiments reveal a clear difference in the classwise performance of both methods, as illustrated in Fig.~\ref{fig:motivation}.
The right half of the figure shows the action classes where the temporally-invariant model is dominant. We can observe that all such action classes are atomic actions with high repetitions, \eg, \texttt{Fencing}, \texttt{Knitting}. Any two clips from such videos are highly similar, hence, increasing agreement between them i.e. learning \textit{temporal-invariant} representation is more meaningful. The left half of the figure shows action classes where temporally-distinctive representations perform better. We can observe that such action classes are slightly more complex i.e. they contain sub-actions, \eg, \texttt{JavelinThrow} first involves running and then throwing. Any two clips from such videos are visually very different, hence if we maximize agreement between them then it results in loss of the temporal dynamics. Therefore, \textit{temporally-distinctive} representation is more suitable in such videos.

Based on our observation, we aim to leverage the strengths of both temporally-invariant and temporally-distinctive representations for semi-supervised action recognition. To achieve this, we propose a semi-supervised framework based on a student-teacher setup. The teacher supervision includes two models pre-trained using CSL with temporally-invariant and temporally-distinctive objectives. After pre-training, the teachers are fine-tuned with the labeled set to adapt to the semi-supervised training of the student. During semi-supervised training, we weigh each teacher model based on the nature of the unlabeled video instance. We determine the nature of the instance by computing its similarity score using the temporal self-similarity matrices of both teachers. This way, the student is trained using the labeled supervision from the labeled set and the unlabeled supervision from the weighted average of the teachers. It is worth noting that our framework doesn't depend on complicated data-augmentation schemes like FixMatch~\cite{fixmatch}.

\noindent{The contributions of this work are summarized as follows:}
\setlist{nolistsep}
\begin{itemize}

\item We propose a student-teacher-based semi-supervised learning framework that consists of two teachers with complementary self-supervised video representations: Temporally-Invariant and Temporally-Distinctive. 

\item In order to leverage the strength of each representation (invariant or distinctive), we weigh a suitable teacher according to an unlabeled video instance. We achieve it by the proposed temporal-similarity-based reweighting scheme. 
\item Our method outperforms existing approaches and achieves state-of-the-art results on popular action recognition benchmarks, including UCF101, HMDB51, and Kinetics400.

\end{itemize}

\section{Prior Work}
\noindent \textbf{Semi-supervised learning in images}
Semi-supervised learning is one of the fundamental approaches to learning from limited labeled data~\cite{Gammerman1998Learning,joachims1999transductive,liu2019deep,kingma2014semi,pu2016variational,simclrv2,rizve2022openldn}. There are two common approaches for semi-supervised learning are consistency regularization \cite{NIPS2016_6333,LaineA17,Miyato2018VirtualAT,tarvainen2017mean, xie2020unsupervised} and pseudo-labeling \cite{Lee2013PseudoLabelT,Shi_2018_ECCV,arazo2020pseudo,ups,zhang2021flexmatch,rizve2022towards}.  The consistency regularization methods attempt to achieve perturbation invariant output space. To this end, these methods try to minimize a distance/divergence-based loss as a measure of consistency between two differently perturbed/augmented versions of an image. Pseudo-labeling-based methods on the other hand promote entropy minimization to improve by performing self-training on the confident pseudo-labels generated from the network. Hybrid methods~\cite{NIPS2019_8749_MixMatch,Berthelot2020ReMixMatch:,fixmatch} combine consistency regularization and pseudo-labeling techniques to obtain further improvement. Some of the recent works~\cite{assran2021semi, s4l, simclrv2} have also shown the effectiveness of self-supervised representations in solving this task. Our work is somewhat in the spirit of this last set of works as we also leverage self-supervised representation learning to obtain temporally distinct and invariant features.

\noindent \textbf{Semi-supervised learning in videos}
Although there is a tremendous recent development in action recognition~\cite{actionreco_bulat2021space, actionreco_chen2022mm, spact, actionreco_mvit, actionreco_mvit2, actionreco_long2022stand, actionreco_ryoo2021tokenlearner, actionreco_timesformer, actionreco_vidtr, c3d,kenshohara,actionreco_arnab2021vivit, actionreco_wang2022long}, semi-supervised learning in videos is not explored as in the image domain. 
CMPL~\cite{semi_cmpl} utilizes a FixMatch framework, where they study the effect of model capacity to provide complementary gains of unlabeled videos. They provide empirical evidence that the smaller model is responsible for learning the motion features whereas, the bigger model learns more appearance-biased features. However, defining smaller and bigger models is very relative, and the observation may not hold true for different architectures. Some prior work injects temporal dynamics to their semi-supervised framework by additional input modality to RGB videos like temporal gradients~\cite{semi_tgfixmatch}, optical-flow~\cite{semi_mvpl}, or P-frames~\cite{semi_compressed}. Another set of methods utilizes consistency loss between the slow and fast input streams of the video to leverage the unlabeled videos~\cite{semi_tacl, semi_tcl}. We can see that the prior works rely on hard-inductive biases like model architectures, input modality, or input sampling to learn temporal-invariance and distinctiveness. On the other hand, we do not have such a hard design choice; we leverage the unlabeled videos by the nature of the video instance using two complementary self-supervised teachers.

\noindent \textbf{Self-supervised learning (SSL)}
In recent years, self-supervised learning has demonstrated learning powerful representations for images~\cite{simclr, moco, swav, zbontar2021barlow, mae, gupta2022higher} and videos~\cite{jenni2021time, tclr, Feichtenhofer_2021_CVPR, cvrl, simon, videomae, stmae, Schiappa_2022}. Although some works~\cite{simclrv2, s4l} have exploited self-supervised representation in semi-supervised image classification, Video SSL is not explored yet for semi-supervised action recognition. 

Video SSL methods can be grouped mainly into two categories: the first set of methods focuses on learning temporal invariance~\cite{cvrl, Feichtenhofer_2021_CVPR, videomoco}.
These methods are a simple-yet-effective extension of instance-discimination-based methods like SimCLR~\cite{simclr}, MoCo~\cite{moco}, etc., where mutual information between two views of an instance is maximized. For videos, two views are two clips from different timestamps, hence it introduces temporal invariance in the learned representations. 
The second set of methods focuses on learning the temporal distinctiveness, where they try to learn different representations for different clips through contrastive loss~\cite{tclr, dsm, seco, iic} or through different temporal pretext transformations~\cite{jenni2021time, simon, rspnet, varpsp, pace_pred, speedNet}.

TCLR~\cite{tclr} introduces temporal contrastive losses for both temporally pooled and unpooled features to learn the temporal distinctiveness. In our method, we utilized these losses to learn the temporal distinctive teacher.

\begin{figure}
    \centering
    \includegraphics[width=\columnwidth]{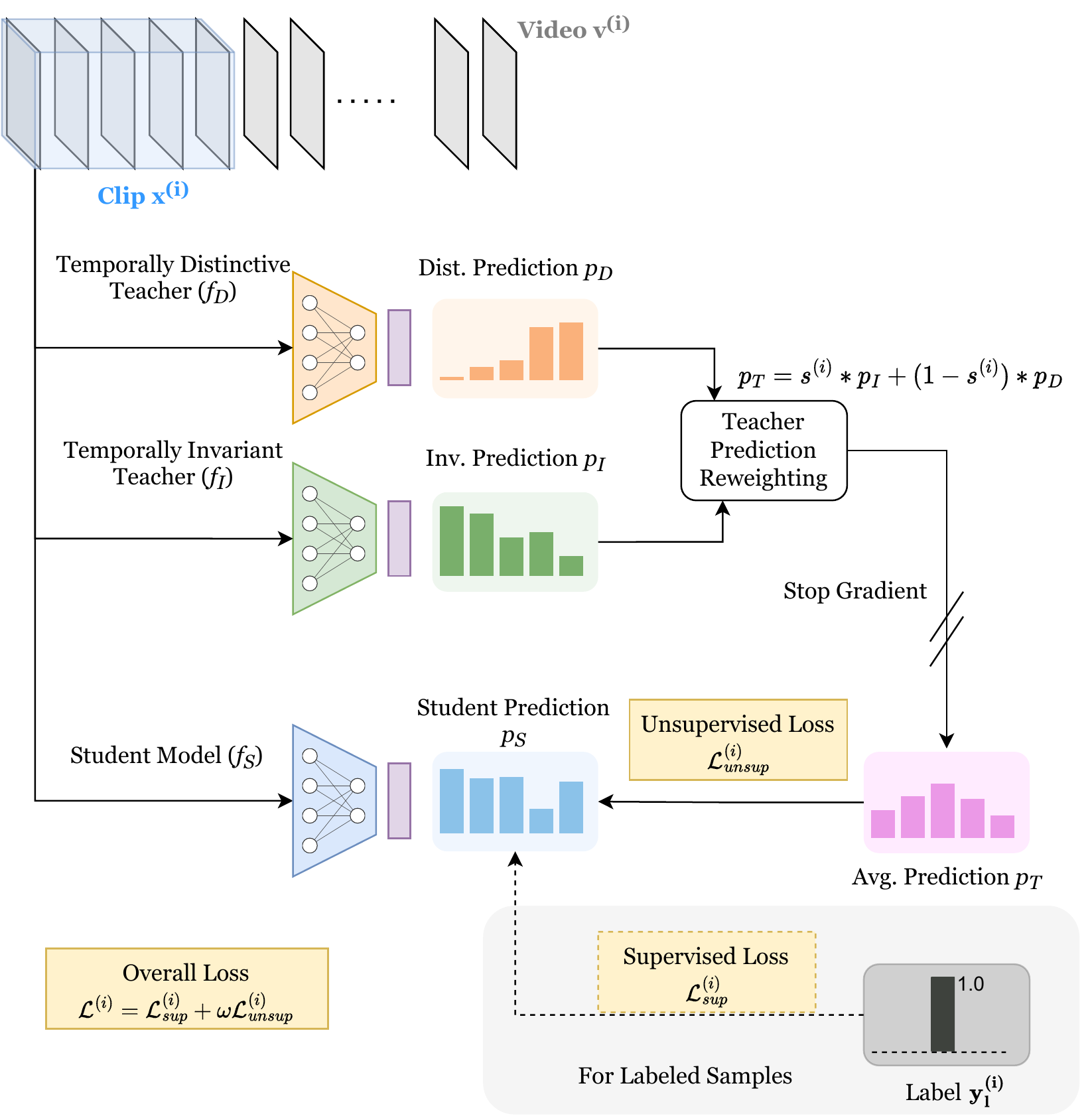}
    \caption{\textbf{Our Framework} We use a teacher-student framework where we use two teachers: $f_{I}$ and $f_{D}$. The input clip $\mathbf{x}^{(i)}$ is given to the teachers and student to get their predictions.
    We utilize a reweighting strategy to combine the predictions of two teachers. Regardless of whether the video $\mathbf{v}^{(i)}$ is labeled or unlabeled, we distill the combined knowledge of teachers to the student. For the labeled samples, we also apply standard cross-entropy loss.}
    \label{fig:framework}
\end{figure}

\section{Method}
Let's consider a small labeled set of videos $\mathbb{D}_{l} = \{(\mathbf{v}^{(i)}, \mathbf{y}^{(i)})\}_{i=1}^{N_{l}}$, where $\mathbf{v}^{(i)}$ and $\mathbf{y}^{(i)}$ denote $i$th video instance and its associated action label and $N_{l}$ is number of total instances in the dataset. We also have access to a unlabeled dataset $\mathbb{D}_{u} = \{\mathbf{v}^{(i)}\}_{i=1}^{N_{u}}$, where $N_u$ is the total number of unlabeled videos and $N_u \gg N_l$. The objective of the semi-supervised action recognition is to utilize both labeled and unlabeled sets ($\mathbb{D}_{l}$ and $\mathbb{D}_{u}$) to improve the action recognition performance.

A high-level schematic diagram of our framework is depicted in Fig.~\ref{fig:framework}. Our semi-supervised learning framework, \textit{TimeBalance}, is a teacher-student framework. To train on the unlabeled samples, we distill the knowledge of two teacher models: temporally-invariant teacher $f_{I}$ and temporally-distinctive teacher $f_{D}$, which are trained in self-supervised manner,  to a student model $f_{S}$.  In the following, we explain the details of \textit{TimeBalance}. To be particular, in Sec.~\ref{sec:sslteachers}, we present the self-supervised pretraining of teacher models. 
In Sec.~\ref{sec:semisup} we explain the semi-supervised training of the student model and our \underline{T}emporal \underline{S}imilarity based \underline{T}eacher \underline{R}eweighting (TSTR) scheme.

\subsection{Self-supervised pretraining of teachers}
\label{sec:sslteachers}
From a video instance $\mathbf{v}^{(i)}$, we sample $n$ consecutive clips $\mathbb{X}^{(i)} = \{\mathbf{x}^{(i)}_{t}\}_{t=1}^{n}$, where $t$ represents clip location (timestamp). Each of these clips undergoes stochastic transformations (e.g. random crop, random color jittering, etc.). Next, we send these clips
to the teacher model $f$ and a non-linear projection head $g$ respectively. The non-linear projection head, projects the clip-embedding from the teacher model to a lower-dimensional normalized representation $\mathbf{z}$, s.t.  $\mathbf{z} \in \mathbf{R}^{d}$, where $d$ is the dimension of output vector $\mathbf{z}$.
\subsubsection{Pretraining of Temporally-Invariant Teacher}
The goal of temporal-invariant pretraining is to learn the shared information across the $n$ different clips $\{\mathbf{x}^{(i)}_{t}\}_{t=1}^{n}$ of the same video instance $i$. 
To achieve this, we maximize the agreement between the projections, $\mathbf{z}$, of two different clips from the same video instance $\{ (\mathbf{z}^{(i)}_{t_1},\mathbf{z}^{(i)}_{t_2}) \mid t_1,t_2 \in \{1...n\} \;and\; t_1 \neq t_2 \}$, while maximizing the disagreement between the projections of clips from different video instances $\{(\mathbf{z}^{(i)},\mathbf{z}^{(j)}) \mid i, j \in B  \;and\; i\neq j\}$, where $B$ is the batch-size. This contrastive objective can be expressed as the following equation:
\begin{equation}\label{eq:inv}
  \mathcal{L}_{I}^{(i)}=- \sum_{\substack{t_1,t_2 \\ t_2\neq t_1}}^{n} \log \frac{\mathrm{h}\left(\mathbf{z}^{(i)}_{t_1}, \mathbf{z}^{(i)}_{t_2}\right)}{\sum\limits_{j=1}^{B}\mathbb{1}_{[j\neq i]} \mathrm{h}(\mathbf{z}^{(i)}_{t_1}, \mathbf{z}^{(j)}_{t_1}) + \mathrm{h}(\mathbf{z}^{(i)}_{t_1}, \mathbf{z}^{(j)}_{t_2})},
\end{equation}
\normalsize	
\noindent where $\mathrm{h}(\mathbf{u_{1}}, \mathbf{u_{2}})=\exp \left(\mathbf{u_{1}}^{T}\mathbf{u_{2}}/(\|\mathbf{u_{1}}\| \|\mathbf{u_{2}}\| \tau) \right)$ is used to compute the similarity between $\mathbf{u_{1}}$ and $\mathbf{u_{2}}$ vectors with an adjustable temperature parameter, $\tau$. $\mathbb{1}_{[j\neq i]} \in \{0, 1\}$ is an indicator function which equals 1 iff $j \neq i$.

\begin{equation}\label{eq:total}
\mathcal{L}^{(i)}= \mathcal{L}^{(i)}_{sup} + \omega \mathcal{L}^{(i)}_{unsup}, 
\end{equation}
\noindent where $\omega$ is the weight of the unsupervised loss.

\begin{figure}
    \centering
    \includegraphics[width=\columnwidth]{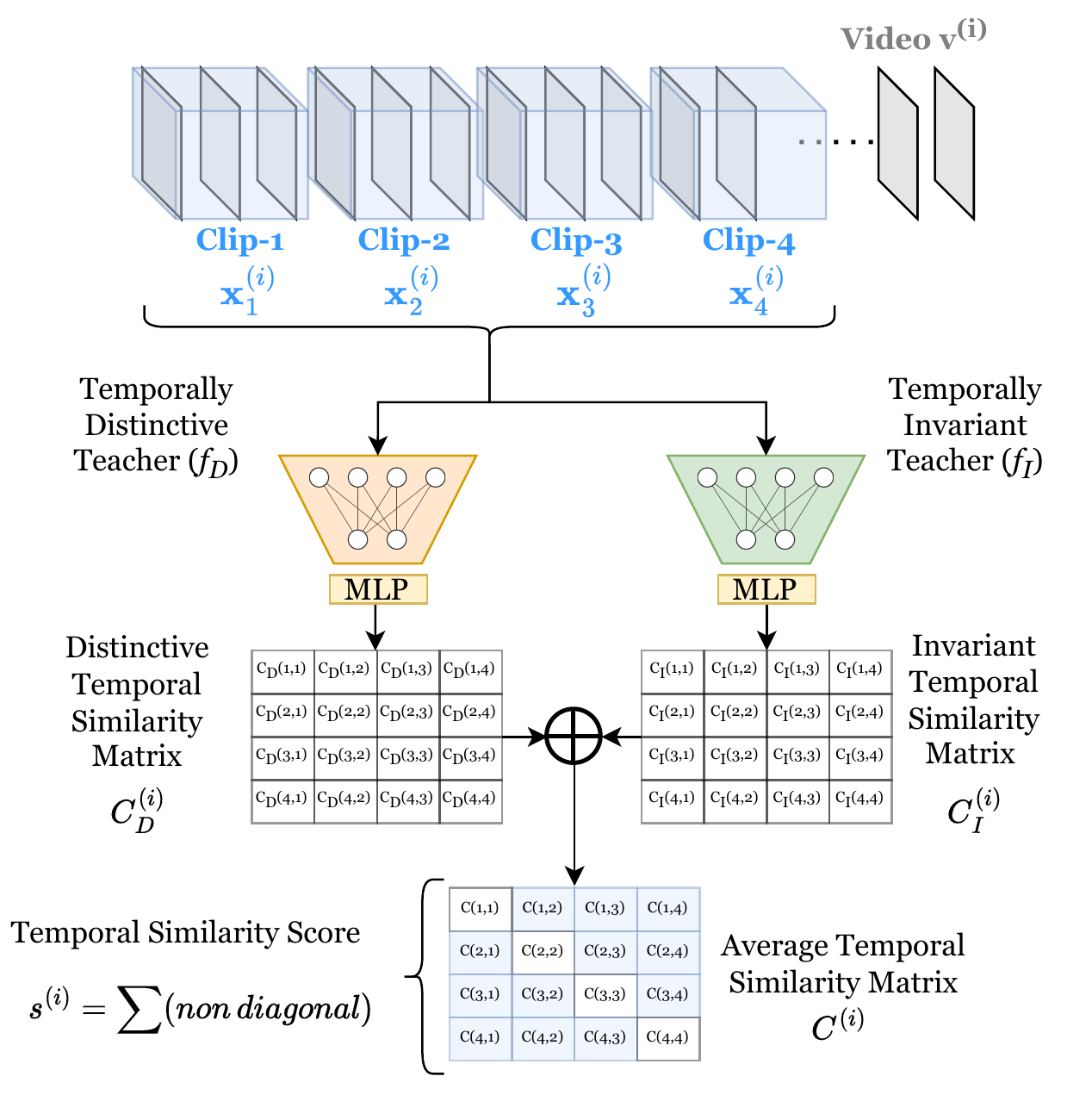}
    \caption{\textbf{Temporal Similarity based Teacher Reweighting} Firstly, a set of clips from video $\mathbf{v}^{(i)}$ are passed through the distinctive and invariant teachers to get representations. Secondly, Temporal Similarity Matrices ($\mathbf{C}^{(i)}_{D}$ and $\mathbf{C}^{(i)}_{I}$) are constructed from the cosine similarity of one timestamp to another timestamp. Finally, from the average matrix $\mathbf{C}^{(i)}$, a temporal similarity score $s^{(i)}$ is computed. $s^{(i)}$ is utilized to combine predictions of teachers for video $\mathbf{v}^{(i)}$ during semi-supervised training.}
    \label{fig:similarity}
\end{figure}

\subsubsection{Pretraining of Temporally-Distinctive Teacher}
Contrary to the goal of $\mathcal{L}_{I}$, temporally-distinctive pretraining deals with learning the differences across the clips of the same video instance. To achieve this, we generate another set of clips $\mathbb{\tilde{X}}^{(i)} = \{\mathbf{\tilde{x}}^{(i)}_{t}\}_{t=1}^{n}$, which are randomly-augmented versions of clips in $\mathbb{X}^{(i)}$. After that, we maximize the agreement between the projections of a pair of clips $\{ (\mathbf{z}^{(i)}_{t_1},\mathbf{\tilde{z}}^{(i)}_{t_1}) \mid t_1 \in \{1..n\} \}$ from the same timestamp and maximize the disagreement between the projections of pair of temporally misaligned clips. A mathematical expression for this contrastive objective can be written as:

\begin{equation}\label{eq:dist}
  \mathcal{L}_{D1}^{(i)}=- \sum_{t_1=1}^{n} \log \frac{\mathrm{h}\left(\mathbf{z}^{(i)}_{t_1}, \mathbf{\tilde{z}}^{(i)}_{t_1}\right)}{\sum\limits_{\substack{t_2=1 \\ t_2\neq t_1}}^{n} \mathrm{h}(\mathbf{z}^{(i)}_{t_1}, \mathbf{z}^{(i)}_{t_2}) + \mathrm{h}(\mathbf{z}^{(i)}_{t_1}, \mathbf{\tilde{z}}^{(i)}_{t_2})},
\end{equation}
\normalsize	

The above contrastive loss imposes temporal distinctiveness at the clip-level i.e. temporally-average-pooled features. Similarly, we can also impose temporal distinctiveness ($\mathcal{L}_{D2}^{(i)}$) on a more fine-grained level i.e. on the unpooled temporal feature slices~\cite{tclr}. More details in \supp{Supp. Sec.~\ref{sec:method_semi_supp}}. We combine these pooled and unpooled temporal-distinctiveness objectives to obtain $\mathcal{L}_{D}^{(i)}=  \mathcal{L}_{D1}^{(i)} + \mathcal{L}_{D2}^{(i)}$.

The primary objective of this work is semi-supervised action recognition. Therefore, even though the self-supervised teacher models lack any explicit notion of category-specific output space, we argue that to solve the downstream action recognition task their knowledge has to be distilled from the action category-specific output space. To this end, we finetune both of the self-supervised teacher models on the labeled set $\mathbb{D}_{l}$ using cross-entropy loss.

\subsection{Semi-supervised training of student model}
\label{sec:semisup}
We initialize the student model with weights from a video self-supervised model~\cite{tclr} trained on $\mathbb{D}_{u}$. 
During the semi-supervised training, student model $f_{S}$ gets supervision from two sources: (i) ground-truth label (if available), and (ii) teacher supervision (Fig.~\ref{fig:framework}). A visual aid is provided in \supp{Supp. Sec.~\ref{sec:method_semi_supp}} for loss computations over the labeled and unlabeled split in semi-supervised training.  The supervision from the ground-truth label is utilized using standard cross-entropy loss, as depicted in the following equation:

\begin{equation}\label{eq:crossentropy}
\mathcal{L}^{(i)}_{sup} = -\sum_{c=1}^{C} \mathbf{y}^{(i)}_{c}\log \mathbf{p}^{(i)}_{c}
\end{equation}

\noindent For instance $i$, the prediction vectors of the invariant and distinctive teacher are denoted as $\mathbf{p}_{I}$ and $\mathbf{p}_{D}$, respectively.

Next, we will discuss our TSTR scheme to distill knowledge from the temporally invariant and distinct teachers to train the student model on the unlabeled data, $\mathbb{D}_u$ and videos of labeled data $\mathbb{D}_l$.

\paragraph{Temporal Similarity based Teacher Reweighting.}
In order to combine supervision from $f_I$ and $f_D$ for a particular video instance $\mathbf{v}^{(i)}$, we first compute temporal similarity scores. To this end, we compute the cosine similarity between each pair of clips to form a temporal similarity matrix, $\mathbf{C}^{(i)}$, as depicted in Fig.~\ref{fig:similarity}. The temporal similarity matrix computation is described in the following,

\begin{equation}\label{eq:similarity1}
\begin{aligned}
\mathbf{C}^{(i)} = \Big[\mathrm{Sim}(\mathbf{z}_{t_1}^{(i)}, \mathbf{z}_{t_2}^{(i)})\Big]_{t_1, t_2=1}^{n},\\
\end{aligned}
\end{equation}
\noindent where, $\mathrm{Sim}(.)$ is the cosine similarity function.

We compute the temporal similarity matrix for both invariant and distinctive teachers denoted as $\mathbf{C}^{(i)}_{I}$, and $\mathbf{C}^{(i)}_{D}$, respectively. Next, in order to get a similarity score, $s^{(i)}$, for an instance $i$, we take the average of non-diagonal elements of both $\mathbf{C}^{(i)}_{I}$, and $\mathbf{C}^{(i)}_{D}$ matrices. 

\begin{equation}\label{eq:similarity2}
s^{(i)} = \frac{1}{2n(n-1)} \sum \limits_{\substack{t_1,t_2=1 \\ t_2\neq t_1}}^{n} (\mathbf{C}_{I}^{(i)} + \mathbf{C}_{D}^{(i)})
\end{equation}

We use this temporal similarity score, $s^{(i)}$, to aggregate the outputs of the teacher models. Let's assume, for instance $i$, the prediction vectors of the invariant and distinctive teacher are denoted as $\mathbf{p}^{(i)}_{I}$ and $\mathbf{p}^{(i)}_{D}$, respectively. Now, we want to combine these teacher prediction vectors in such a way that the temporally-invariant prediction $\mathbf{p}^{(i)}_{I}$ gets a higher weight if the temporal similarity score is high, and the weight of the temporal-distinctive prediction $\mathbf{p}^{(i)}_{D}$ gets higher in the case of a lower temporal similarity score. This dynamic weighting scheme for obtaining the aggregated teacher prediction is provided below.  

\begin{equation}\label{eq:reweighting}
\mathbf{p}^{(i)}_{T} = s^{(i)}.\mathbf{p}^{(i)}_{I} + (1-s^{(i)}).\mathbf{p}^{(i)}_{D}
\end{equation}

We use the combined teacher prediction $\mathbf{p}^{(i)}_{T}$ to provide supervision to the student model using $\mathcal{L}_2$ loss. We compute this loss only on the unlabeled samples ($\mathbb{D}_u$ and videos of $\mathbb{D}_l$ without assigned labels), hence, we refer to this loss as the unsupervised loss. This loss term is defined below.

\begin{equation}\label{eq:l2}
\mathcal{L}^{(i)}_{unsup} =\left(\mathbf{p}^{(i)}_T-\mathbf{p}^{(i)}_S\right)^2
\end{equation}

Finally, we sum both the supervised and unsupervised losses to train the student model. The overall objective function is defined below.

\subsection{Algorithm}
Let's consider models $f_{I}$, $f_{D}$, and $f_{S}$ are parameterized by $\theta_{I}$, $\theta_{D}$, and $\theta_{S}$.  All steps of our semi-supervised training are put together in Algorithm~\ref{alg:algo1}.

\begin{algorithm}
\textbf{Inputs}:
            
        \hspace*{0.2cm} \textit{Datasets:} $\mathbb{D}_{u}$, $\mathbb{D}_{l}$\\
        \hspace*{0.2cm} \textit{\#Epochs:} $max\_ssl\_epoch_{I}$, $max\_ssl\_epoch_{D}$, $max\_epoch_{tune}$, 
        $max\_epoch$\\
        \hspace*{0.2cm} \textit{Learning Rates: $\alpha_I$, $\alpha_D$, $\alpha_S$}
        
\textbf{Output}: Student model $\theta_{S}$

\SetAlgoLined
Initialize $\theta_I$, $\theta_D$ randomly;

Initialize $\theta_S$ with any SSL~\cite{tclr, cvrl} method on $\mathbb{D}_{u}$;

\hrule

Temporally-Invariant Self-supervised Pretraining:

\For{$e_0 \gets 1$ \KwTo $max\_ssl\_epoch_{I}$}
{   
    $\theta_{I} \gets \theta_I-\alpha_I\nabla_{\theta_I}L_{I}(\theta_I)$
}
Temporally-Distinctive Self-supervised Pretraining:

\For{$e_0 \gets 1$ \KwTo $max\_ssl\_epoch_{D}$}
{   
    $\theta_{D} \gets \theta_D-\alpha_D\nabla_{\theta_D}L_{D}(\theta_D)$
}

Compute the similarity score $s^{(i)}$ using Eq.~\ref{eq:similarity1}
\hrule

Finetuning the teacher models on labeled set $\mathbb{D}_{l}$:

\For{$e_0 \gets 1$ \KwTo $max\_epoch_{tune}$}
{   
    $\theta_{I}^{*} = argmin_{\theta_{I}} L_{sup}(\theta_{I})$ 
    
    $\theta_{D}^{*} = argmin_{\theta_{D}} L_{sup}(\theta_{D})$
    
}

\hrule

Semi-supervised training of student on $\mathbb{D}_{l}$ + $\mathbb{D}_{u}$:

\For{$e_0 \gets 1$ \KwTo $max\_epoch$}
{   

    $L$= $L_{sup}(\theta_S) +  \omega L_{unsup}(\theta_S, \theta_{D}^{*}, \theta_{I}^{*}, s^{(i)})$
    
    $\theta_{S} \gets \theta_S- \alpha_S\nabla_{\theta_S}L$

}
\caption{TimeBalance training algorithm}
 \label{alg:algo1}

\end{algorithm}

\begin{table*}[h]
\centering
\arrayrulecolor[rgb]{0.8,0.8,0.8}
\small
\begingroup
\setlength{\tabcolsep}{3pt}
\arrayrulecolor[rgb]{0.753,0.753,0.753}
\begin{tabular}{llccc!{\color{black}\vrule}c|c|c|c|c!{\color{black}\vrule}c|c|c!{\color{black}\vrule}c|c} 
\arrayrulecolor{black}\hline

\hline

\hline\\[-3mm]
\multirow{2}{*}{\textbf{Method}}       & \multirow{2}{*}{\textbf{Backbone}} & \multirow{2}{*}{\begin{tabular}[c]{@{}c@{}}\textbf{Params}\\\textbf{(M)}\end{tabular}} & \multirow{2}{*}{\textbf{Input}} & \multirow{2}{*}{\textbf{\#F}} & \multicolumn{5}{c!{\color{black}\vrule}}{\textbf{UCF101}}                   & \multicolumn{3}{c!{\color{black}\vrule}}{\textbf{HMDB51}} & \multicolumn{2}{c}{\textbf{Kinetics400}}  \\ 
\cline{6-15}
                                       &                                    &                                                                                        &                                 &                               & \textbf{1\%} & \textbf{5\%} & \textbf{10\%} & \textbf{20\%} & \textbf{50\%} & \textbf{40\%} & \textbf{50\%} & \textbf{60\%}               & \textbf{1\%} & \textbf{10\%}              \\ 
\hline
PL~\venue{ICML'13}~\cite{lee2013pseudo}                             & 3D-ResNet18                        & 13.5                                                                                   & V                               & 16                            & -            & 17.6         & 24.7          & 37.0          & 47.5          & 27.3        & 32.4          & 33.5                        & -            & -                          \\ 
MT~\venueTT{NeuRIPS'17}~\cite{tarvainen2017mean}                         & 3D-ResNet18                        & 13.5                                                                                   & V                               & 16                            & -            & 17.5         & 25.6          & 36.3          & 45.8          & 27.2        & 30.4          & 32.2                        & -            & -                          \\
S4L~\venueTT{ICCV'19}~\cite{s4l}                            & 3D-ResNet18                        & 13.5                                                                                   & V                               & 16                            & -            & 22.7         & 29.1          & 37.7          & 47.9          & 29.8        & 31.0          & 35.6                        & -            & -                          \\
UPS~\venueTT{ICLR'21}~\cite{ups}                            & 3D-ResNet18                        & 13.5                                                                                   & V                               & 16                            & -            & -            & -             & 39.4          & 50.2          & -           & -             & -                           & -            & -                          \\
SD~\venueTT{ICCV'19}~\cite{girdhar2019distinit}                             & 3D-ResNet18                        & 13.5                                                                                   & V                               & 16                            & -            & 31.2         & 40.7          & 45.4          & 53.9          & 32.6        & 35.1          & 36.3                        & -            & -                          \\
MT+SD ~\venueTT{WACV'21}~\cite{semi_videossl}                       & 3D-ResNet18                        & 13.5                                                                                   & V                               & 16                            & -            & 30.3         & 40.5          & 45.5          & 53.0          & 32.3        & 33.6          & 35.7                        & -            & -                          \\
3DRotNet~\venueTT{Arxiv'19}~\cite{3drotnet}                          & 3D-ResNet18                        & 13.5                                                                                   & V                               & 16                            & 15.0            & 31.5        & 40.4          & 47.1          & -         & -        & -          & -                        & -            & -                          \\
\textcolor[rgb]{0.6,0.6,0.6}{VideoSemi~\venueTT{WACV'21}~\cite{semi_videossl}} & \textcolor[rgb]{0.6,0.6,0.6}{3D-ResNet18} & \textcolor[rgb]{0.6,0.6,0.6}{13.5}                                               & \textcolor[rgb]{0.6,0.6,0.6}{V} & \textcolor[rgb]{0.6,0.6,0.6}{16} & \textcolor[rgb]{0.6,0.6,0.6}{-} & \textcolor[rgb]{0.6,0.6,0.6}{32.4} & \textcolor[rgb]{0.6,0.6,0.6}{42.0} & \textcolor[rgb]{0.6,0.6,0.6}{48.7} & \textcolor[rgb]{0.6,0.6,0.6}{54.3} & \textcolor[rgb]{0.6,0.6,0.6}{32.7} & \textcolor[rgb]{0.6,0.6,0.6}{36.2} & \textcolor[rgb]{0.6,0.6,0.6}{37.0} & \textcolor[rgb]{0.6,0.6,0.6}{-} & \textcolor[rgb]{0.6,0.6,0.6}{-}  \\
TCL~\venueTT{CVPR'21}~\cite{semi_tcl}                            & TSM-ResNet18                       & -                                                                                      & V                               & 8                             & -            & -            & -             & -             & -             & -           & -             & -                           & 11.6         & -                          \\
TG-FixMatch~\venueTT{CVPR'21}~\cite{semi_tgfixmatch}                    & 3D-ResNet18                        & 13.5                                                                                   & V                               & 8                             & -            & \secondbest{44.8}         & {62.4}          & \secondbest{76.1}          & \secondbest{79.3}          & \secondbest{46.5}        & \secondbest{48.4}          & \secondbest{49.7}                        & 9.8          & 43.8                       \\
MvPL~\venueTT{ICCV'21}~\cite{semi_mvpl}                           & 3D-ResNet18                        & 13.5                                                                                   & VFG                             & 8                             & -            & 41.2         & 55.5          & 64.7          & 65.6          & 30.5        & 33.9          & 35.8                        & 5.0          & 36.9                       \\
TCLR~\venueTT{CVIU'22}~\cite{tclr}                           & 3D-ResNet18                        & 13.5                                                                                   & V                               & 16                            & \secondbest{26.9}         & -            & 66.1          & 73.4          & 76.7          & -           & -             & -                           & -            & -                          \\
CMPL~\venueTT{CVPR'22}~\cite{semi_cmpl}                           & 3D-ResNet18                        & 13.5                                                                                   & V                               & 8                             & 23.8         & -            & \secondbest{67.6}          & -             & -             & -           & -             & -                           & \secondbest{16.5}         & \secondbest{53.7}                       \\
TACL~\venueTT{TSVT'22}~\cite{semi_tacl}                           & 3D-ResNet18                        & 13.5                                                                                   & V                               & 16                            & -            & 35.6         & 50.9          & 56.1          & 65.8          & 34.6        & 37.2          & 39.5                        & -            & -                          \\
\textcolor[rgb]{0.6,0.6,0.6}{TACL~\venueTT{TSVT'22}~\cite{semi_tacl}}         & \textcolor[rgb]{0.6,0.6,0.6}{3D-ResNet18} & \textcolor[rgb]{0.6,0.6,0.6}{13.5}                                               & \textcolor[rgb]{0.6,0.6,0.6}{V} & \textcolor[rgb]{0.6,0.6,0.6}{16} & \textcolor[rgb]{0.6,0.6,0.6}{-} & \textcolor[rgb]{0.6,0.6,0.6}{43.7} & \textcolor[rgb]{0.6,0.6,0.6}{55.6} & \textcolor[rgb]{0.6,0.6,0.6}{59.2} & \textcolor[rgb]{0.6,0.6,0.6}{67.2} & \textcolor[rgb]{0.6,0.6,0.6}{38.7} & \textcolor[rgb]{0.6,0.6,0.6}{40.2} & \textcolor[rgb]{0.6,0.6,0.6}{41.7} & \textcolor[rgb]{0.6,0.6,0.6}{-} & \textcolor[rgb]{0.6,0.6,0.6}{-}  \\
MemDPC~\venueTT{ECCV'20}~\cite{memdpc}                           & 3D-ResNet18                        & 13.5                                                                                   & V                               & 16                            & -            & -         & 44.2          & 50.9          & 62.3          & -        & -          & -                        & -            & -                          \\
MotionFit~\venueTT{ICCV'21}~\cite{motionfit}                           & 3D-ResNet18                        & 13.5                                                                                   & VF                               & 16                            & -            & -         & -          & 57.7          & 59.0          & -        & -         & -                        & -            & -                          \\

\rowcolor[rgb]{0.784,0.902,0.976} Ours \textit{(TimeBalance)} & 3D-ResNet18                        & 13.5                                                                                   & V                               & 8                             & \bestresult{29.1}         & \bestresult{47.9}         & \bestresult{69.8}          & \bestresult{79.1}          & \bestresult{83.3}          & \bestresult{49.8}        & \bestresult{51.4}          & \bestresult{53.1}                        & \bestresult{17.1}         & \bestresult{54.9}                       \\ 
\arrayrulecolor{black}\hline
ActorCM~\venueTT{Arxiv'21}~\cite{zou2021learning}                   & R(2+1)D-34                         & 33.3                                                                                   & V                               & 8                             & -            & \secondbest{27.0}         & 40.2          & \secondbest{51.7}          & \secondbest{59.9}          & \secondbest{32.9}        & \secondbest{38.2}          & \secondbest{38.9}                        & -            & -                          \\ 
\textcolor[rgb]{0.6,0.6,0.6}{ActorCM~\venueTT{Arxiv'21}~\cite{zou2021learning}} & \textcolor[rgb]{0.6,0.6,0.6}{R(2+1)D-34}  & \textcolor[rgb]{0.6,0.6,0.6}{33.3}                                               & \textcolor[rgb]{0.6,0.6,0.6}{V} & \textcolor[rgb]{0.6,0.6,0.6}{8}  & \textcolor[rgb]{0.6,0.6,0.6}{-} & \textcolor[rgb]{0.6,0.6,0.6}{45.1} & \textcolor[rgb]{0.6,0.6,0.6}{53.0} & \textcolor[rgb]{0.6,0.6,0.6}{57.4} & \textcolor[rgb]{0.6,0.6,0.6}{64.7} & \textcolor[rgb]{0.6,0.6,0.6}{35.7} & \textcolor[rgb]{0.6,0.6,0.6}{39.5} & \textcolor[rgb]{0.6,0.6,0.6}{40.8} & \textcolor[rgb]{0.6,0.6,0.6}{-} & \textcolor[rgb]{0.6,0.6,0.6}{-}  \\
FixMatch~\venueTT{NeuRIPS'20}~\cite{fixmatch}                    & SlowFast-R50                       & 60                                                                                     & V                               & 8                             & 16.1         & -            & 55.1          & -             & -             & -           & -             & -                           & 10.1         & 49.4                       \\
MvPL~\venueTT{ICCV'21}\cite{semi_mvpl}                           & 3D-ResNet50                        & 31.8                                                                                   & VFG                             & 8                             & 22.8         & -            & \secondbest{80.5}          & -             & -             & -           & -             & -                           & 17.0         & 58.2                       \\
CMPL~\venueTT{CVPR'22}~\cite{semi_cmpl}                           & 3D-ResNet50                        & 31.8                                                                                   & V                               & 8                             & \secondbest{25.1}         & -            & 79.1          & -             & -             & -           & -             & -                           & \secondbest{17.6}         & \secondbest{58.4}                       \\
\rowcolor[rgb]{0.784,0.902,0.976} Ours \textit{(TimeBalance)} & 3D-ResNet50                        & 31.8                                                                                   & V                               & 8                             & \bestresult{30.1}         & \bestresult{53.5}         & \bestresult{81.1}          & \bestresult{83.3}          & \bestresult{85.0}          & \bestresult{52.6}        & \bestresult{53.9}          & \bestresult{54.5}                        & \bestresult{19.6}         & \bestresult{61.2}                       \\
\arrayrulecolor{black}\hline
\end{tabular}
\endgroup
\caption{\textbf{Comparison with state-of-the-art methods}: Methods using pre-trained ImageNet weights are shown in \textcolor[rgb]{0.6,0.6,0.6}{Grey}. V- Video (RGB), F- Optical Flow, G- Temporal Gradients. Best results are shown in \bestresult{Red}, and Second-best in \secondbest{Blue}}
\label{table:big}

\end{table*}

\section{Experiments}
\subsection{Datasets}
\noindent\textbf{UCF101}~\cite{ucf101} is a dataset for human action recognition collected from internet videos consisting 101 action classes. We use split-1 for our experiments, which has 9,537 train videos and 3,783 test videos.

\noindent\textbf{HMDB51}~\cite{hmdb} is relatively a smaller dataset collected from movie videos. It has 51 human activity classes and has a high intra-class variance. We use split-1 in this paper, which has 3,570 train videos and 1,530 test videos.

\noindent \textbf{Kinetics400}~\cite{kinetics} is a large-scale dataset collected from YouTube videos. It has a standard split of 240k training videos and 20k validation videos which covers 400 actions.

\subsection{Implementation Details}
For our default experimental setup, we use an input clip resolution of $224\times224$. We use 3D-ResNet-50~\cite{slowfast} for both student and teacher models. We use most commonly used augmentations: geometry-based augmentations like random cropping, and flipping, and color-based augmentations like random greyscale, and color jittering.

\noindent \textbf{Self-supervised pretraining} 
The pretraining is performed with clips of $16$ frames for 100 epochs for Kinetics400, 250 epochs for UCF101, and HMDB51 experiments. For contrastive losses, the default temperature is set to 0.1.

\noindent \textbf{Semi-supervised training} 
Semi-supervised training is performed with clips of $8$ frames for 150 epochs. 

\noindent \textbf{Inference} 
We follow standard protocol~\cite{r2plus1d} of averaging the predictions of 10 uniformly spaced clips and 3 different spatial scales to get a video-level prediction.
\\
More implementation details in \supp{Supp. Sec.~\ref{sec:impl_semi_supp}}. 

\begin{figure*}
    
    \begin{subfigure}{0.49\textwidth}
        \centering
        \includegraphics[clip, trim=0cm 3.2cm 0cm 0cm, width=0.99\textwidth]{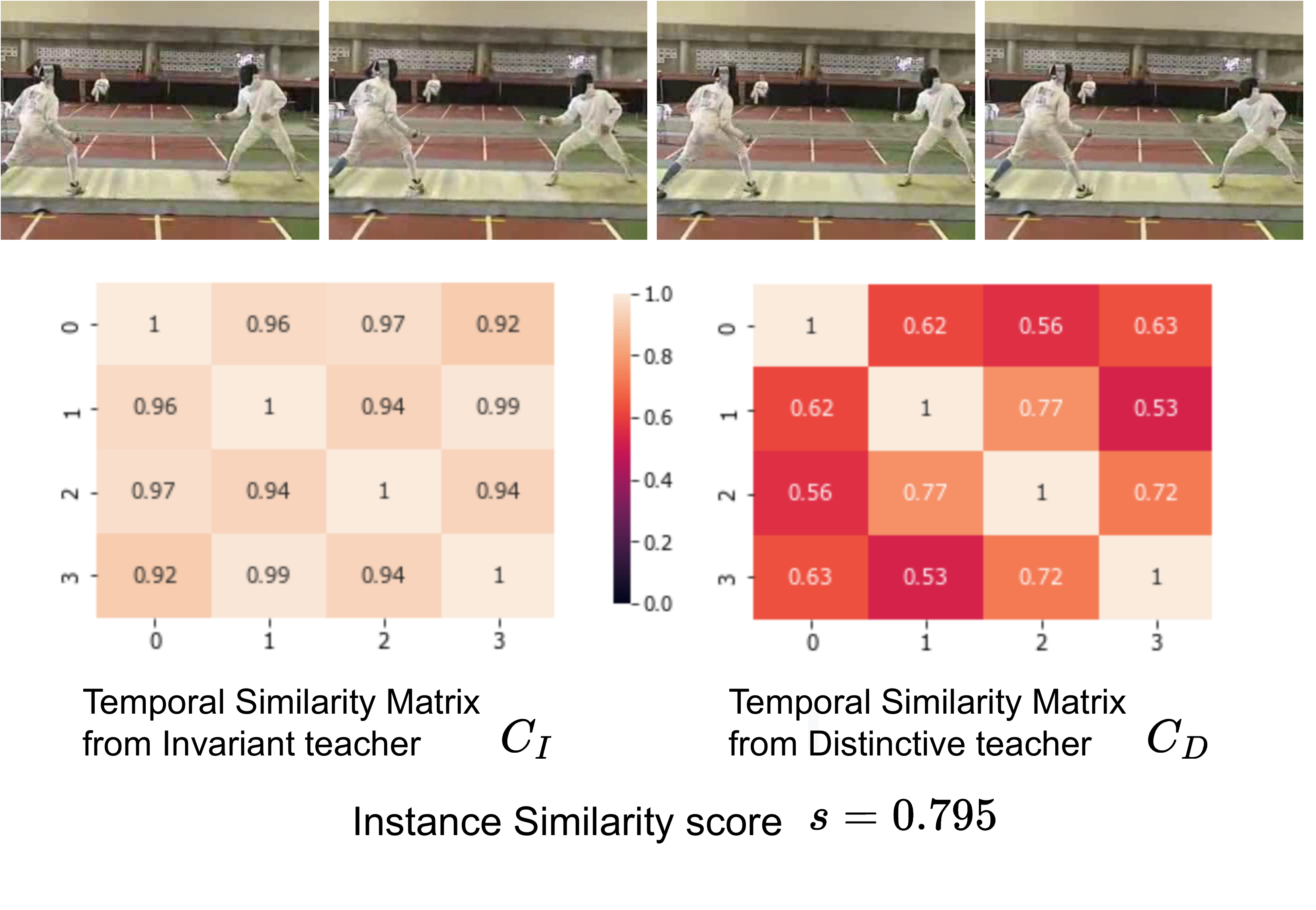}
        \vspace{-0.5mm}
        \caption{\texttt{Fencing}, $s^{(i)}=0.795$}

    \end{subfigure}
    \begin{subfigure}{0.49\textwidth}
        \centering
        \includegraphics[clip, trim=0cm 3.2cm 0cm 0cm, width=0.99\textwidth]{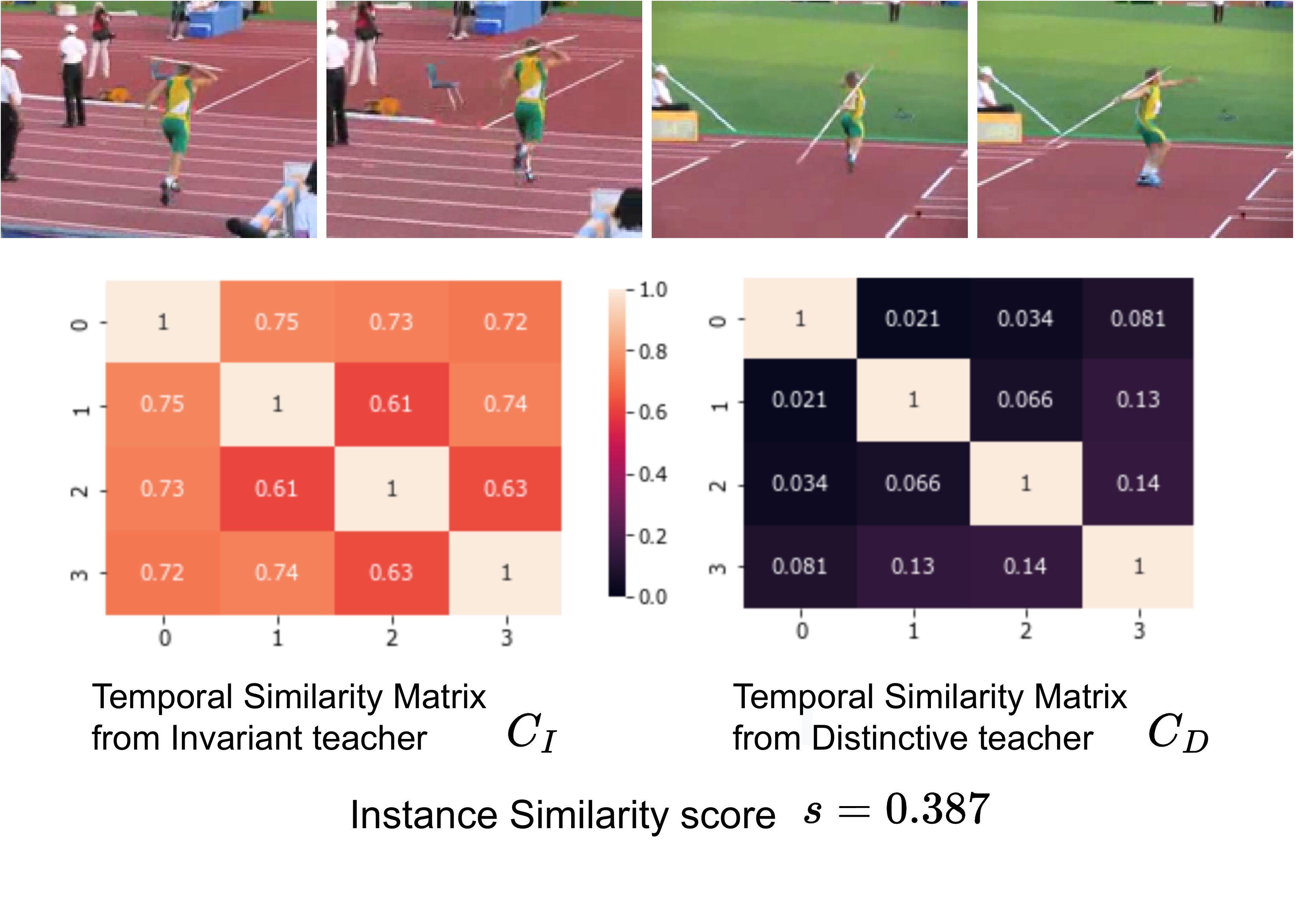}
        \vspace{-0.5mm}
        \caption{\texttt{JavelinThrow}, $s^{(i)}=0.387$}

    \end{subfigure}

    \caption{Visualization of similarity matrices from temporally-invariant and distinctive teachers and resultant instance-similarity score. (a) Video instances with atomic actions like \texttt{Fencing} result in a high instance similarity score, which will proportionally increase the weightage of temporally-invariant teacher (b) Whereas, complex actions video instances like  \texttt{JavelinThrow} results in a low similarity score, which results into more weightage of temporally-distinctive teacher. Details of the computation of the similarity score in Sec~\ref{sec:semisup} }
    \label{fig:similarity_matrix_visualize}
\end{figure*}

\subsection{Comparison with prior work}
We compare our method with image-based baselines, video self-supervised baselines, and recent state-of-the-art methods for semi-supervised action recognition in Table~\ref{table:big}. We present the results in two sections based on the backbone architecture used: (1) 3D-ResNet18 and (2) 3D-ResNet50. We report Top-1 classification accuracy as the performance measure and follow the standard protocol of reporting average performance over three independent runs.

\noindent \textbf{Image-based baselines} We consider widely used semi-supervised image classification baselines like Psuedo-Label~\cite{lee2013pseudo}, MeanTeacher~\cite{tarvainen2017mean}, UPS~\cite{ups}, S4L~\cite{s4l} and FixMatch~\cite{fixmatch}. From Table~\ref{table:big}, we observe that the results of these image-based methods are significantly lower than the video-based methods across all benchmarks; which suggests that spatial information is not enough to excel in semi-supervised action recognition.

\noindent \textbf{Video Self-supervised Learning baselines} Video self-supervised learning methods are first pretrained on the full-training set without using any labels and then finetuned on the labeled set in a supervised manner. We compare with contrastive learning-based methods like TCLR~\cite{tclr}, MemDPC~\cite{memdpc}, multimodal approach like MotionFit~\cite{motionfit}, and pretext-task based approach 3DRotNet~\cite{3drotnet}.  From Table~\ref{table:big}, we notice that the performance of these video self-supervised methods is significantly better than image-based baselines, and in some cases, it even performs favorably against the semi-supervised action recognition methods~\cite{semi_cmpl, semi_mvpl, semi_tgfixmatch}. However, our proposed semi-supervised method outperforms all these video self-supervised baselines by a noticeable margin. 

\noindent \textbf{Semi-supervised action recognition baselines} We also compare with prior semi-supervised action recognition works like TG-Fixmatch~\cite{semi_tgfixmatch}, VideoSemi~\cite{semi_videossl}, MvPL~\cite{semi_mvpl}, TACL~\cite{semi_tacl}, and ActorCM~\cite{semi_actorcutmix}. Our method achieves significant improvement over these methods across all benchmarks. Remarkably, our method also outperforms the methods that use additional modalities like optical flow in MvPL~\cite{semi_mvpl} and additional data (ImageNet) in ActorCM~\cite{semi_actorcutmix} and TACL~\cite{semi_tacl}.

\subsection{Ablations and Analysis}
\label{sec:ablation}
In the default setting of ablations, we consider UCF101 with 3D-ResNet50 as the teacher backbone. More ablations in \supp{Supp. Sec.~\ref{sec:ablation_semi_supp}}. 

\noindent \textbf{Contribution of different training components}
In Table~\ref{table:abl_framework}, we analyze the effect of each model ($f_S$, $f_I$, $f_D$) and our proposed teacher reweighting scheme with 5\% and 20\% labeled data on the UCF101 dataset. \texttt{Row a-c}, demonstrates the performance of teacher and student models individually. Results in \texttt{Row f} demonstrates that the student model performs the best when we average the predictions of both teachers and removing any of them(\texttt{Row d, e}) degrades the performance. This validates our hypothesis that for optimal performance, we need to distill the knowledge from both teachers. Finally, \texttt{Row g}, demonstrates the effectiveness of our proposed teacher reweighting scheme using temporal similarity. We found similar results with $f_S$ from random initialization in Table~\ref{table:abl_scratch_framework}.

\begin{table}[h]
\centering
\begingroup
\setlength{\tabcolsep}{4pt}
\begin{tabular}{lcccccc} 
\arrayrulecolor{black}\hline

\hline

\hline\\[-3mm]
& \multirow{2}{*}{\textbf{$f_{S}$}} & \multirow{2}{*}{\textbf{$f_{I}$}} & \multirow{2}{*}{\textbf{$f_{D}$}} & \multicolumn{1}{l}{\multirow{2}{*}{\begin{tabular}[c]{@{}l@{}}\textbf{Teacher}\\\textbf{Reweighting}\end{tabular}}} & \multicolumn{2}{c}{\textbf{UCF101 \% labels}}  \\
 &                            &                              &                              &                                                                                                 & \textbf{5\%} & \textbf{20\%}                   \\ 
\hline
\texttt{(a)} & \cmark                           & \textcolor[rgb]{0.7,0.7,0.7}{\xmark}                           & \textcolor[rgb]{0.7,0.7,0.7}{\xmark}                           & \textcolor[rgb]{0.7,0.7,0.7}{\xmark}                                                                                              & 48.66        & 79.70                           \\
\texttt{(b)} & \textcolor[rgb]{0.7,0.7,0.7}{\xmark}                           & \cmark                           & \textcolor[rgb]{0.7,0.7,0.7}{\xmark}                           & \textcolor[rgb]{0.7,0.7,0.7}{\xmark}                                                                                              & 43.79        & 74.55                           \\
\texttt{(c)} & \textcolor[rgb]{0.7,0.7,0.7}{\xmark}                           & \textcolor[rgb]{0.7,0.7,0.7}{\xmark}                           & \cmark                           & \textcolor[rgb]{0.7,0.7,0.7}{\xmark}                                                                                              & 44.08        & 75.13                           \\
\hline
\texttt{(d)} & \cmark                           & \cmark                           & \textcolor[rgb]{0.7,0.7,0.7}{\xmark}                           & \textcolor[rgb]{0.7,0.7,0.7}{\xmark}                                                                                              & 47.95        & 79.28                           \\
\texttt{(e)} & \cmark                           & \textcolor[rgb]{0.7,0.7,0.7}{\xmark}                           & \cmark                           & \textcolor[rgb]{0.7,0.7,0.7}{\xmark}                                                                                              & 48.81        & 79.76                           \\
\texttt{(f)} & \cmark                           & \cmark                           & \cmark                           & \textcolor[rgb]{0.7,0.7,0.7}{\xmark}                                                                                              & 52.14        & 82.02                           \\ 
\hline
\texttt{(g)} & \cmark                           & \cmark                           & \cmark                           & \cmark                                                                                              & 53.48        & 83.26                           \\
\arrayrulecolor{black}\hline

\hline

\hline\\[-3mm]
\end{tabular}

\endgroup

\caption{Ablation of different components of our framework. Student ($f_{S}$) and teachers ($f_{I}$ and $f_{D}$) are 3D-ResNet50.}

\label{table:abl_framework}
\end{table}
\begin{table}[h]
\centering
\begingroup
\setlength{\tabcolsep}{1.2pt}
\begin{tabular}{ccclccc} 
\arrayrulecolor{black}\hline

\hline

\hline\\[-3mm]
 & \multirow{2}{*}{\begin{tabular}[c]{@{}l@{}}\textbf{$f_{S}$}\\\textbf{(rand. init.)}\end{tabular}} & \multirow{2}{*}{\textbf{$f_{I}$}} & \multirow{2}{*}{\textbf{$f_{D}$}} & \multirow{2}{*}{\begin{tabular}[c]{@{}l@{}}\textbf{Teacher}\\\textbf{Reweighting}\end{tabular}} & \multicolumn{2}{c}{\textbf{UCF101 \% Labels}}  \\
                                                                                             &                              &                              &                                                                                       &          & \textbf{5\%} & \textbf{20\%}                   \\ 
\hline
\texttt{(a)} & \cmark                                                                                           & \textcolor[rgb]{0.7,0.7,0.7}{\xmark}                           & \textcolor[rgb]{0.7,0.7,0.7}{\xmark}                           & \textcolor[rgb]{0.7,0.7,0.7}{\xmark}                                                                                              & 25.60        & 46.20                           \\
\texttt{(b)} & \cmark                                                                                           & \cmark                           & \textcolor[rgb]{0.7,0.7,0.7}{\xmark}                           & \textcolor[rgb]{0.7,0.7,0.7}{\xmark}                                                                                              & 43.94        & 74.85                           \\
\texttt{(c)} & \cmark                                                                                           & \textcolor[rgb]{0.7,0.7,0.7}{\xmark}                           & \cmark                           & \textcolor[rgb]{0.7,0.7,0.7}{\xmark}                                                                                              & 44.30        & 75.22                           \\
\texttt{(d)} & \cmark                                                                                           & \cmark                           & \cmark                           & \textcolor[rgb]{0.7,0.7,0.7}{\xmark}                                                                                              & 49.57        & 80.06                           \\
\hline
\texttt{(e)} & \cmark                                                                                           & \cmark                           & \cmark                           & \cmark                                                                                              & 53.10        & 83.00                           \\
\arrayrulecolor{black}\hline

\hline

\hline\\[-3mm]
\end{tabular}

\endgroup
\caption{Ablation of different components with Student with random initialization. All backbones are 3D-ResNet50.}

\label{table:abl_scratch_framework}
\end{table}
\noindent \textbf{Visualization of Temporal Similarity Matrix}
We sample four consecutive frames from unlabeled videos and compute the temporal similarity matrix for both $f_I$ and $f_D$. Visualization is shown Fig.~\ref{fig:similarity_matrix_visualize}. Firstly, we observe that $f_I$ gives higher similarity values in both video instances than $f_D$. Secondly, we can observe that the final instance similarity score aligns with our goal i.e. providing more weight to $f_I$ in atomic actions and providing more weight to $f_D$ in complex actions.

\noindent \textbf{Different types of teacher pairs}
In order to further investigate the importance of having teachers with complementary self-supervised objectives (i.e temporally invariant and distinctive), we study the effect of having various pairs of teachers in our framework (Table~\ref{table:abl_ensemble}). \textit{Inv1} and \textit{Inv2} models are obtained from independent SSL pretrainings with temporal-invariance objective. Similarly, \textit{Dist1} and \textit{Dist2} are trained with temporal-distinctiveness SSL objective. We observe that, compared to combining teachers with the same SSL objective (\texttt{Row a, b}), teachers with different SSL objectives (\texttt{Row c, e}) perform significantly better. In our default setting, we use (\textit{Inv1}, \textit{Dist1}) as teacher pair.

\begin{table}[h]
\centering
\begingroup
\setlength{\tabcolsep}{6pt}
\begin{tabular}{lllcc} 
\arrayrulecolor{black}\hline

\hline

\hline\\[-3mm]
&\multirow{2}{*}{\textbf{Teacher-1}} & \multirow{2}{*}{\textbf{Teacher-2}} & \multicolumn{2}{c}{\textbf{UCF101 \% labels}}  \\
 &                                   &                                     & \textbf{5\%} & \textbf{20\%}                   \\ 
\hline
\texttt{(a)} & Inv1                                & Inv2                                & 48.33        & 78.76                           \\
\texttt{(b)} & Dist1                               & Dist2                               & 49.15        & 80.49                           \\
\texttt{(c)} & Inv1                                & Dist1                               & 52.14        & 82.02                           \\
\texttt{(d)} & Inv1                                & Dist2                               & 51.78        & 81.43                           \\
\arrayrulecolor{black}\hline

\hline

\hline\\[-3mm]
\end{tabular}
\endgroup

\caption{Ablation of different teacher combinations.}

\label{table:abl_ensemble}
\end{table}

\begin{figure}
    \centering
    \includegraphics[trim={5mm 0 2mm 3mm},clip, width=0.8\columnwidth]{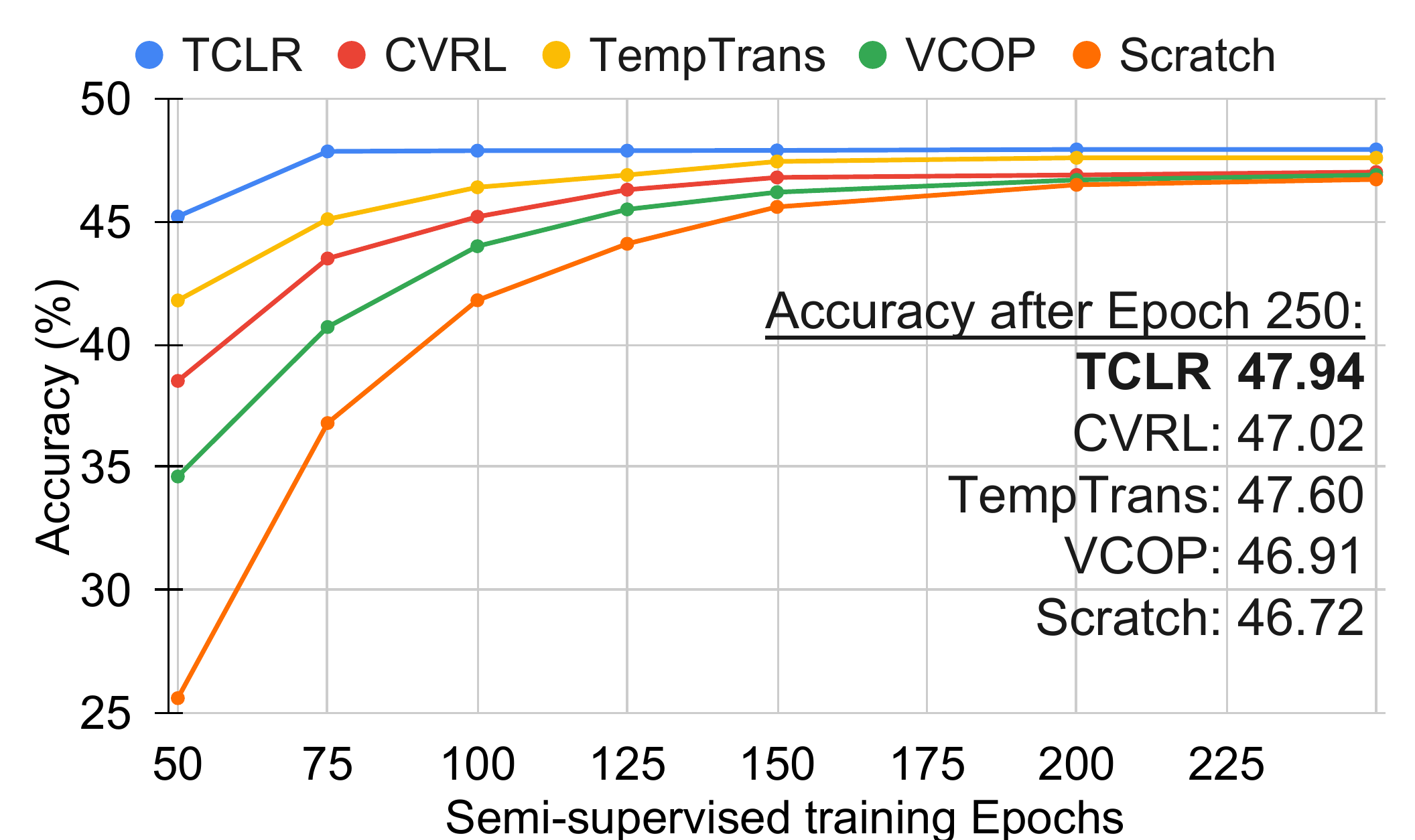}
    \caption{Analysis of different student initialization strategies on UCF101 dataset with 5\% labeled data and  3D-ResNet18 student.}
    \label{fig:epochwise}
\end{figure}

\noindent \textbf{Student initialization}
We also study various video self-supervised methods to initialize the student model and report the results in Fig~\ref{fig:epochwise}.
We conduct these experiments on the UCF101 dataset with 3D-ResNet18 as the student backbone. We observe that even across a diverse set of video SSL-based initialization techniques, our proposed semi-supervised framework can achieve better/competitive performance against the current state-of-the-art. This further validates the effectiveness of our idea of leveraging temporally distinct and invariant teachers and our proposed teacher prediction reweighting scheme. 

\noindent \textbf{Number of clips in Teacher SSL pretraining}
We analyze the effect of using a different number of clips to impose temporal invariance/distinctiveness in teacher pretraining in Table~\ref{table:abl_nclips}. We perform pretraining on UCF101 with the 3D-ResNet50 model and report the performance of the teacher model after finetuning on the labeled set. In the default setting, we use n=4 clips. We observe that while reducing the clips from 4 to 2, the performance drop is significant for temporally-distinctive teachers, whereas, it is not severe for the temporally-invariant teacher. Since the distinctiveness contrastive loss $\mathcal{L}_{D}$ treats other clips as negatives, its contrastive objective becomes very easy to solve if we reduce the number of clips (Eq.~\ref{eq:dist}). Whereas, temporal-invariance contrastive loss $\mathcal{L}_{I}$ takes negatives from the other instances of the batch, and decreasing the number of clips does not change the difficulty of the loss.

\begin{table}[h]
\centering
\begingroup
\setlength{\tabcolsep}{5pt}
\begin{tabular}{llcc} 
\arrayrulecolor{black}\hline

\hline

\hline\\[-3mm]
\multirow{2}{*}{\textbf{Pretraining}} & \multirow{2}{*}{\begin{tabular}[c]{@{}l@{}}\textbf{Number of }\\\textbf{clips (n)}\end{tabular}} & \multicolumn{2}{c}{\textbf{UCF101 \% Labels}}  \\
                                      &                                               & \textbf{5\%} & \textbf{20\%}                   \\ 
\hline
\multirow{2}{*}{Invariant Teacher}    & n=4                                           & 43.79        & 74.55                           \\
                                      & n=2                                           & 42.78        & 73.86                           \\ 
\hline
\multirow{2}{*}{Distinctive Teacher}  & n=4                                           & 44.08        & 75.13                           \\
                                      & n=2                                           & 39.55        & 71.21                           \\
\arrayrulecolor{black}\hline

\hline

\hline\\[-3mm]
\end{tabular}
\endgroup

\caption{Number of clips per video in self-supervised pretraining.}
\label{table:abl_nclips}
\vspace{-2mm}

\end{table}

\section{Conclusion and Future Work}
In this work, we have proposed TimeBalance, a teacher-student framework for semi-supervised action recognition. We utilize the complementary strengths of temporally-invariant and temporally-distinctive representations to leverage unlabeled videos. Our extensive experimentation has empirically validated the effectiveness of different components of our framework and results on multiple benchmark datasets established TimeBalance as the new state-of-the-art for semi-supervised action recognition. 

Our findings regarding the complementary strengths of temporally-invariant and temporally-distinctive video representations could be applied to other data-efficient video understanding problems, such as few-shot action recognition and spatio-temporal action detection. It would also be interesting to explore the invariance and distinctiveness properties of video using recent masked-image modeling techniques~\cite{videomae, stmae} and multi-modal settings~\cite{bachmann2022multimae}, such as audio + video and text + video.

\section*{Acknowledgments}
\noindent We thank Tristan de Blegiers for his help with visualization.

\clearpage
{\small
\bibliographystyle{ieee_fullname}
\bibliography{main}

\begin{thebibliography}{10}\itemsep=-1pt

\bibitem{arazo2020pseudo}
Eric Arazo, Diego Ortego, Paul Albert, Noel~E O’Connor, and Kevin McGuinness.
\newblock Pseudo-labeling and confirmation bias in deep semi-supervised
  learning.
\newblock In {\em 2020 International Joint Conference on Neural Networks
  (IJCNN)}, pages 1--8. IEEE, 2020.

\bibitem{actionreco_arnab2021vivit}
Anurag Arnab, Mostafa Dehghani, Georg Heigold, Chen Sun, Mario Lu{\v{c}}i{\'c},
  and Cordelia Schmid.
\newblock Vivit: A video vision transformer.
\newblock In {\em Proceedings of the IEEE/CVF International Conference on
  Computer Vision}, pages 6836--6846, 2021.

\bibitem{assran2021semi}
Mahmoud Assran, Mathilde Caron, Ishan Misra, Piotr Bojanowski, Armand Joulin,
  Nicolas Ballas, and Michael Rabbat.
\newblock Semi-supervised learning of visual features by non-parametrically
  predicting view assignments with support samples.
\newblock In {\em Proceedings of the IEEE/CVF International Conference on
  Computer Vision}, pages 8443--8452, 2021.

\bibitem{bachmann2022multimae}
Roman Bachmann, David Mizrahi, Andrei Atanov, and Amir Zamir.
\newblock {MultiMAE}: Multi-modal multi-task masked autoencoders.
\newblock 2022.

\bibitem{speedNet}
Sagie Benaim, Ariel Ephrat, Oran Lang, Inbar Mosseri, William~T Freeman,
  Michael Rubinstein, Michal Irani, and Tali Dekel.
\newblock Speednet: Learning the speediness in videos.
\newblock In {\em Proceedings of the IEEE/CVF Conference on Computer Vision and
  Pattern Recognition}, pages 9922--9931, 2020.

\bibitem{actionreco_timesformer}
Gedas Bertasius, Heng Wang, and Lorenzo Torresani.
\newblock Is space-time attention all you need for video understanding?
\newblock In {\em ICML}, volume~2, page~4, 2021.

\bibitem{Berthelot2020ReMixMatch:}
David Berthelot, Nicholas Carlini, Ekin~D. Cubuk, Alex Kurakin, Kihyuk Sohn,
  Han Zhang, and Colin Raffel.
\newblock Remixmatch: Semi-supervised learning with distribution matching and
  augmentation anchoring.
\newblock In {\em International Conference on Learning Representations}, 2020.

\bibitem{NIPS2019_8749_MixMatch}
David Berthelot, Nicholas Carlini, Ian Goodfellow, Nicolas Papernot, Avital
  Oliver, and Colin~A Raffel.
\newblock Mixmatch: A holistic approach to semi-supervised learning.
\newblock In {\em Advances in Neural Information Processing Systems 32}, pages
  5049--5059. Curran Associates, Inc., 2019.

\bibitem{actionreco_bulat2021space}
Adrian Bulat, Juan~Manuel Perez~Rua, Swathikiran Sudhakaran, Brais Martinez,
  and Georgios Tzimiropoulos.
\newblock Space-time mixing attention for video transformer.
\newblock {\em Advances in Neural Information Processing Systems},
  34:19594--19607, 2021.

\bibitem{buzzelli2020vision}
Marco Buzzelli, Alessio Alb{\'e}, and Gianluigi Ciocca.
\newblock A vision-based system for monitoring elderly people at home.
\newblock {\em Applied Sciences}, 10(1):374, 2020.

\bibitem{swav}
Mathilde Caron, Ishan Misra, Julien Mairal, Priya Goyal, Piotr Bojanowski, and
  Armand Joulin.
\newblock Unsupervised learning of visual features by contrasting cluster
  assignments.
\newblock In H. Larochelle, M. Ranzato, R. Hadsell, M.~F. Balcan, and H. Lin,
  editors, {\em Advances in Neural Information Processing Systems}, volume~33,
  pages 9912--9924. Curran Associates, Inc., 2020.

\bibitem{kinetics}
Joao Carreira and Andrew Zisserman.
\newblock Quo vadis, action recognition? a new model and the kinetics dataset.
\newblock In {\em proceedings of the IEEE Conference on Computer Vision and
  Pattern Recognition}, pages 6299--6308, 2017.

\bibitem{actionreco_chen2022mm}
Jiawei Chen and Chiu~Man Ho.
\newblock Mm-vit: Multi-modal video transformer for compressed video action
  recognition.
\newblock In {\em Proceedings of the IEEE/CVF Winter Conference on Applications
  of Computer Vision}, pages 1910--1921, 2022.

\bibitem{rspnet}
Peihao Chen, Deng Huang, Dongliang He, Xiang Long, Runhao Zeng, Shilei Wen,
  Mingkui Tan, and Chuang Gan.
\newblock Rspnet: Relative speed perception for unsupervised video
  representation learning.
\newblock In {\em The AAAI Conference on Artificial Intelligence (AAAI)}, 2021.

\bibitem{simclr}
Ting Chen, Simon Kornblith, Mohammad Norouzi, and Geoffrey Hinton.
\newblock A simple framework for contrastive learning of visual
  representations.
\newblock In {\em ICML}, 2020.

\bibitem{simclrv2}
Ting Chen, Simon Kornblith, Kevin Swersky, Mohammad Norouzi, and Geoffrey~E
  Hinton.
\newblock Big self-supervised models are strong semi-supervised learners.
\newblock {\em Advances in Neural Information Processing Systems}, 33, 2020.

\bibitem{varpsp}
Hyeon Cho, Taehoon Kim, Hyung~Jin Chang, and Wonjun Hwang.
\newblock Self-supervised visual learning by variable playback speeds
  prediction of a video.
\newblock {\em IEEE Access}, 9:79562--79571, 2021.

\bibitem{tclr}
Ishan Dave, Rohit Gupta, Mamshad~Nayeem Rizve, and Mubarak Shah.
\newblock Tclr: Temporal contrastive learning for video representation.
\newblock {\em Computer Vision and Image Understanding}, page 103406, 2022.

\bibitem{gabv2}
Ishan Dave, Zacchaeus Scheffer, Akash Kumar, Sarah Shiraz, Yogesh~Singh Rawat,
  and Mubarak Shah.
\newblock Gabriellav2: Towards better generalization in surveillance videos for
  action detection.
\newblock In {\em Proceedings of the IEEE/CVF Winter Conference on Applications
  of Computer Vision (WACV) Workshops}, pages 122--132, January 2022.

\bibitem{spact}
Ishan~Rajendrakumar Dave, Chen Chen, and Mubarak Shah.
\newblock Spact: Self-supervised privacy preservation for action recognition.
\newblock In {\em Proceedings of the IEEE/CVF Conference on Computer Vision and
  Pattern Recognition}, pages 20164--20173, 2022.

\bibitem{hvu}
Ali Diba, Mohsen Fayyaz, Vivek Sharma, Manohar Paluri, J{\"u}rgen Gall, Rainer
  Stiefelhagen, and Luc Van~Gool.
\newblock Large scale holistic video understanding.
\newblock In {\em European Conference on Computer Vision}, pages 593--610.
  Springer, 2020.

\bibitem{actionreco_mvit}
Haoqi Fan, Bo Xiong, Karttikeya Mangalam, Yanghao Li, Zhicheng Yan, Jitendra
  Malik, and Christoph Feichtenhofer.
\newblock Multiscale vision transformers.
\newblock In {\em Proceedings of the IEEE/CVF International Conference on
  Computer Vision}, pages 6824--6835, 2021.

\bibitem{stmae}
Christoph Feichtenhofer, Haoqi Fan, Yanghao Li, and Kaiming He.
\newblock Masked autoencoders as spatiotemporal learners.
\newblock In {\em Advances in Neural Information Processing Systems}, 2022.

\bibitem{slowfast}
Christoph Feichtenhofer, Haoqi Fan, Jitendra Malik, and Kaiming He.
\newblock Slowfast networks for video recognition.
\newblock In {\em Proceedings of the IEEE international conference on computer
  vision}, pages 6202--6211, 2019.

\bibitem{Feichtenhofer_2021_CVPR}
Christoph Feichtenhofer, Haoqi Fan, Bo Xiong, Ross Girshick, and Kaiming He.
\newblock A large-scale study on unsupervised spatiotemporal representation
  learning.
\newblock In {\em Proceedings of the IEEE/CVF Conference on Computer Vision and
  Pattern Recognition (CVPR)}, pages 3299--3309, June 2021.

\bibitem{Gammerman1998Learning}
A. Gammerman, V. Vovk, and V. Vapnik.
\newblock Learning by transduction.
\newblock In {\em Proceedings of the Fourteenth Conference on Uncertainty in
  Artificial Intelligence}, UAI'98, page 148–155, San Francisco, CA, USA,
  1998. Morgan Kaufmann Publishers Inc.

\bibitem{motionfit}
Kirill Gavrilyuk, Mihir Jain, Ilia Karmanov, and Cees~GM Snoek.
\newblock Motion-augmented self-training for video recognition at smaller
  scale.
\newblock In {\em Proceedings of the IEEE/CVF International Conference on
  Computer Vision}, pages 10429--10438, 2021.

\bibitem{girdhar2019distinit}
Rohit Girdhar, Du Tran, Lorenzo Torresani, and Deva Ramanan.
\newblock Distinit: Learning video representations without a single labeled
  video.
\newblock In {\em Proceedings of the IEEE/CVF International Conference on
  Computer Vision}, pages 852--861, 2019.

\bibitem{byol}
Jean-Bastien Grill, Florian Strub, Florent Altch{\'e}, Corentin Tallec, Pierre
  Richemond, Elena Buchatskaya, Carl Doersch, Bernardo Avila~Pires, Zhaohan
  Guo, Mohammad Gheshlaghi~Azar, et~al.
\newblock Bootstrap your own latent-a new approach to self-supervised learning.
\newblock {\em Advances in Neural Information Processing Systems},
  33:21271--21284, 2020.

\bibitem{gupta2022higher}
Rohit Gupta, Naveed Akhtar, Ajmal Mian, and Mubarak Shah.
\newblock On higher adversarial susceptibility of contrastive self-supervised
  learning.
\newblock {\em arXiv preprint arXiv:2207.10862}, 2022.

\bibitem{memdpc}
Tengda Han, Weidi Xie, and Andrew Zisserman.
\newblock Memory-augmented dense predictive coding for video representation
  learning.
\newblock In {\em Computer Vision--ECCV 2020: 16th European Conference,
  Glasgow, UK, August 23--28, 2020, Proceedings, Part III 16}, pages 312--329.
  Springer, 2020.

\bibitem{kenshohara}
K. {Hara}, H. {Kataoka}, and Y. {Satoh}.
\newblock Towards good practice for action recognition with spatiotemporal 3d
  convolutions.
\newblock In {\em 2018 24th International Conference on Pattern Recognition
  (ICPR)}, pages 2516--2521, 2018.

\bibitem{mae}
Kaiming He, Xinlei Chen, Saining Xie, Yanghao Li, Piotr Doll{\'a}r, and Ross
  Girshick.
\newblock Masked autoencoders are scalable vision learners.
\newblock In {\em Proceedings of the IEEE/CVF Conference on Computer Vision and
  Pattern Recognition}, pages 16000--16009, 2022.

\bibitem{moco}
Kaiming He, Haoqi Fan, Yuxin Wu, Saining Xie, and Ross Girshick.
\newblock Momentum contrast for unsupervised visual representation learning.
\newblock In {\em Proceedings of the IEEE/CVF Conference on Computer Vision and
  Pattern Recognition}, pages 9729--9738, 2020.

\bibitem{behaviour}
Neziha Jaouedi, Noureddine Boujnah, Oumayma Htiwich, and Med~Salim Bouhlel.
\newblock Human action recognition to human behavior analysis.
\newblock In {\em 2016 7th International Conference on Sciences of Electronics,
  Technologies of Information and Telecommunications (SETIT)}, pages 263--266.
  IEEE, 2016.

\bibitem{jenni2021time}
Simon Jenni and Hailin Jin.
\newblock Time-equivariant contrastive video representation learning.
\newblock In {\em Proceedings of the IEEE/CVF International Conference on
  Computer Vision}, pages 9970--9980, 2021.

\bibitem{simon}
Simon Jenni, Givi Meishvili, and Paolo Favaro.
\newblock Video representation learning by recognizing temporal
  transformations.
\newblock In {\em The European Conference on Computer Vision (ECCV)}, August
  2020.

\bibitem{semi_videossl}
Longlong Jing, Toufiq Parag, Zhe Wu, Yingli Tian, and Hongcheng Wang.
\newblock Videossl: Semi-supervised learning for video classification.
\newblock In {\em Proceedings of the IEEE/CVF Winter Conference on Applications
  of Computer Vision}, pages 1110--1119, 2021.

\bibitem{3drotnet}
Longlong Jing, Xiaodong Yang, Jingen Liu, and Yingli Tian.
\newblock Self-supervised spatiotemporal feature learning via video rotation
  prediction.
\newblock {\em arXiv preprint arXiv:1811.11387}, 2018.

\bibitem{joachims1999transductive}
Thorsten Joachims.
\newblock Transductive inference for text classification using support vector
  machines.
\newblock In {\em Icml}, volume~99, pages 200--209, 1999.

\bibitem{adam}
Diederik~P. Kingma and Jimmy Ba.
\newblock Adam: {A} method for stochastic optimization.
\newblock In Yoshua Bengio and Yann LeCun, editors, {\em 3rd International
  Conference on Learning Representations, {ICLR} 2015, San Diego, CA, USA, May
  7-9, 2015, Conference Track Proceedings}, 2015.

\bibitem{kingma2014semi}
Durk~P Kingma, Shakir Mohamed, Danilo~Jimenez Rezende, and Max Welling.
\newblock Semi-supervised learning with deep generative models.
\newblock In {\em Advances in neural information processing systems}, pages
  3581--3589, 2014.

\bibitem{hmdb}
H. Kuehne, H. Jhuang, E. Garrote, T. Poggio, and T. Serre.
\newblock {HMDB}: a large video database for human motion recognition.
\newblock In {\em Proceedings of the International Conference on Computer
  Vision (ICCV)}, 2011.

\bibitem{LaineA17}
Samuli Laine and Timo Aila.
\newblock Temporal ensembling for semi-supervised learning.
\newblock In {\em ICLR (Poster)}. OpenReview.net, 2017.

\bibitem{Lee2013PseudoLabelT}
Dong-Hyun Lee.
\newblock Pseudo-label : The simple and efficient semi-supervised learning
  method for deep neural networks.
\newblock 2013.

\bibitem{lee2013pseudo}
Dong-Hyun Lee et~al.
\newblock Pseudo-label: The simple and efficient semi-supervised learning
  method for deep neural networks.
\newblock In {\em Workshop on challenges in representation learning, ICML},
  volume~3, page 896, 2013.

\bibitem{li2021multisports}
Yixuan Li, Lei Chen, Runyu He, Zhenzhi Wang, Gangshan Wu, and Limin Wang.
\newblock Multisports: A multi-person video dataset of spatio-temporally
  localized sports actions.
\newblock In {\em Proceedings of the IEEE/CVF International Conference on
  Computer Vision}, pages 13536--13545, 2021.

\bibitem{actionreco_mvit2}
Yanghao Li, Chao-Yuan Wu, Haoqi Fan, Karttikeya Mangalam, Bo Xiong, Jitendra
  Malik, and Christoph Feichtenhofer.
\newblock Improved multiscale vision transformers for classification and
  detection.
\newblock {\em arXiv preprint arXiv:2112.01526}, 2021.

\bibitem{liu2019deep}
Bin Liu, Zhirong Wu, Han Hu, and Stephen Lin.
\newblock Deep metric transfer for label propagation with limited annotated
  data.
\newblock In {\em Proceedings of the IEEE International Conference on Computer
  Vision Workshops}, pages 0--0, 2019.

\bibitem{liu2020privacy}
Jixin Liu, Rong Tan, Guang Han, Ning Sun, and Sam Kwong.
\newblock Privacy-preserving in-home fall detection using visual shielding
  sensing and private information-embedding.
\newblock {\em IEEE Transactions on Multimedia}, 2020.

\bibitem{cmu2020}
Wenhe Liu, Guoliang Kang, Po-Yao Huang, Xiaojun Chang, Yijun Qian, Junwei
  Liang, Liangke Gui, Jing Wen, and Peng Chen.
\newblock Argus: Efficient activity detection system for extended video
  analysis.
\newblock In {\em Proceedings of the IEEE/CVF Winter Conference on Applications
  of Computer Vision (WACV) Workshops}, March 2020.

\bibitem{actionreco_long2022stand}
Fuchen Long, Zhaofan Qiu, Yingwei Pan, Ting Yao, Jiebo Luo, and Tao Mei.
\newblock Stand-alone inter-frame attention in video models.
\newblock In {\em Proceedings of the IEEE/CVF Conference on Computer Vision and
  Pattern Recognition}, pages 3192--3201, 2022.

\bibitem{Miyato2018VirtualAT}
Takeru Miyato, Shin ichi Maeda, Masanori Koyama, and Shin Ishii.
\newblock Virtual adversarial training: A regularization method for supervised
  and semi-supervised learning.
\newblock {\em IEEE Transactions on Pattern Analysis and Machine Intelligence},
  41:1979--1993, 2018.

\bibitem{videomoco}
Tian Pan, Yibing Song, Tianyu Yang, Wenhao Jiang, and Wei Liu.
\newblock Videomoco: Contrastive video representation learning with temporally
  adversarial examples.
\newblock In {\em Proceedings of the IEEE/CVF Conference on Computer Vision and
  Pattern Recognition}, pages 11205--11214, 2021.

\bibitem{pu2016variational}
Yunchen Pu, Zhe Gan, Ricardo Henao, Xin Yuan, Chunyuan Li, Andrew Stevens, and
  Lawrence Carin.
\newblock Variational autoencoder for deep learning of images, labels and
  captions.
\newblock In {\em Advances in neural information processing systems}, pages
  2352--2360, 2016.

\bibitem{cvrl}
Rui Qian, Tianjian Meng, Boqing Gong, Ming-Hsuan Yang, Huisheng Wang, Serge
  Belongie, and Yin Cui.
\newblock Spatiotemporal contrastive video representation learning.
\newblock In {\em Proceedings of the IEEE/CVF Conference on Computer Vision and
  Pattern Recognition}, pages 6964--6974, 2021.

\bibitem{rizve2021gabriella}
Mamshad~Nayeem Rizve, Ugur Demir, Praveen Tirupattur, Aayush~Jung Rana, Kevin
  Duarte, Ishan~R Dave, Yogesh~S Rawat, and Mubarak Shah.
\newblock Gabriella: An online system for real-time activity detection in
  untrimmed security videos.
\newblock In {\em 2020 25th International Conference on Pattern Recognition
  (ICPR)}, pages 4237--4244. IEEE, 2021.

\bibitem{ups}
Mamshad~Nayeem Rizve, Kevin Duarte, Yogesh~S Rawat, and Mubarak Shah.
\newblock In defense of pseudo-labeling: An uncertainty-aware pseudo-label
  selection framework for semi-supervised learning.
\newblock In {\em International Conference on Learning Representations}, 2021.

\bibitem{rizve2022openldn}
Mamshad~Nayeem Rizve, Navid Kardan, Salman Khan, Fahad Shahbaz~Khan, and
  Mubarak Shah.
\newblock Openldn: Learning to discover novel classes for open-world
  semi-supervised learning.
\newblock In {\em Computer Vision--ECCV 2022: 17th European Conference, Tel
  Aviv, Israel, October 23--27, 2022, Proceedings, Part XXXI}, pages 382--401.
  Springer, 2022.

\bibitem{rizve2022towards}
Mamshad~Nayeem Rizve, Navid Kardan, and Mubarak Shah.
\newblock Towards realistic semi-supervised learning.
\newblock In {\em Computer Vision--ECCV 2022: 17th European Conference, Tel
  Aviv, Israel, October 23--27, 2022, Proceedings, Part XXXI}, pages 437--455.
  Springer, 2022.

\bibitem{actionreco_ryoo2021tokenlearner}
Michael Ryoo, AJ Piergiovanni, Anurag Arnab, Mostafa Dehghani, and Anelia
  Angelova.
\newblock Tokenlearner: Adaptive space-time tokenization for videos.
\newblock {\em Advances in Neural Information Processing Systems},
  34:12786--12797, 2021.

\bibitem{NIPS2016_6333}
Mehdi Sajjadi, Mehran Javanmardi, and Tolga Tasdizen.
\newblock Regularization with stochastic transformations and perturbations for
  deep semi-supervised learning.
\newblock In D.~D. Lee, M. Sugiyama, U.~V. Luxburg, I. Guyon, and R. Garnett,
  editors, {\em Advances in Neural Information Processing Systems 29}, pages
  1163--1171. Curran Associates, Inc., 2016.

\bibitem{Schiappa_2022}
Madeline~C. Schiappa, Yogesh~S. Rawat, and Mubarak Shah.
\newblock Self-supervised learning for videos: A survey.
\newblock {\em {ACM} Computing Surveys}, dec 2022.

\bibitem{Shi_2018_ECCV}
Weiwei Shi, Yihong Gong, Chris Ding, Zhiheng MaXiaoyu~Tao, and Nanning Zheng.
\newblock Transductive semi-supervised deep learning using min-max features.
\newblock In {\em The European Conference on Computer Vision (ECCV)}, September
  2018.

\bibitem{semi_tcl}
Ankit Singh, Omprakash Chakraborty, Ashutosh Varshney, Rameswar Panda, Rogerio
  Feris, Kate Saenko, and Abir Das.
\newblock Semi-supervised action recognition with temporal contrastive
  learning.
\newblock In {\em Proceedings of the IEEE/CVF Conference on Computer Vision and
  Pattern Recognition}, pages 10389--10399, 2021.

\bibitem{fixmatch}
Kihyuk Sohn, David Berthelot, Nicholas Carlini, Zizhao Zhang, Han Zhang,
  Colin~A Raffel, Ekin~Dogus Cubuk, Alexey Kurakin, and Chun-Liang Li.
\newblock Fixmatch: Simplifying semi-supervised learning with consistency and
  confidence.
\newblock {\em Advances in neural information processing systems}, 33:596--608,
  2020.

\bibitem{ucf101}
Khurram Soomro, Amir~Roshan Zamir, and Mubarak Shah.
\newblock Ucf101: A dataset of 101 human actions classes from videos in the
  wild.
\newblock {\em arXiv preprint arXiv:1212.0402}, 2012.

\bibitem{iic}
Li Tao, Xueting Wang, and Toshihiko Yamasaki.
\newblock Self-supervised video representation learning using inter-intra
  contrastive framework.
\newblock In {\em Proceedings of the 28th ACM International Conference on
  Multimedia}, pages 2193--2201, 2020.

\bibitem{tarvainen2017mean}
Antti Tarvainen and Harri Valpola.
\newblock Mean teachers are better role models: Weight-averaged consistency
  targets improve semi-supervised deep learning results.
\newblock {\em Advances in neural information processing systems}, 30, 2017.

\bibitem{semi_compressed}
Hayato Terao, Wataru Noguchi, Hiroyuki Iizuka, and Masahito Yamamoto.
\newblock Compressed video ensemble based pseudo-labeling for semi-supervised
  action recognition.
\newblock {\em Machine Learning with Applications}, page 100336, 2022.

\bibitem{semi_tacl}
Anyang Tong, Chao Tang, and Wenjian Wang.
\newblock Semi-supervised action recognition from temporal augmentation using
  curriculum learning.
\newblock {\em IEEE Transactions on Circuits and Systems for Video Technology},
  2022.

\bibitem{videomae}
Zhan Tong, Yibing Song, Jue Wang, and Limin Wang.
\newblock Video{MAE}: Masked autoencoders are data-efficient learners for
  self-supervised video pre-training.
\newblock In {\em Advances in Neural Information Processing Systems}, 2022.

\bibitem{c3d}
Du Tran, Lubomir Bourdev, Rob Fergus, Lorenzo Torresani, and Manohar Paluri.
\newblock Learning spatiotemporal features with 3d convolutional networks.
\newblock In {\em Proceedings of the IEEE international conference on computer
  vision}, pages 4489--4497, 2015.

\bibitem{r2plus1d}
Du Tran, Heng Wang, Lorenzo Torresani, Jamie Ray, Yann LeCun, and Manohar
  Paluri.
\newblock A closer look at spatiotemporal convolutions for action recognition.
\newblock In {\em Proceedings of the IEEE conference on Computer Vision and
  Pattern Recognition}, pages 6450--6459, 2018.

\bibitem{actionreco_wang2022long}
Jue Wang, Gedas Bertasius, Du Tran, and Lorenzo Torresani.
\newblock Long-short temporal contrastive learning of video transformers.
\newblock In {\em Proceedings of the IEEE/CVF Conference on Computer Vision and
  Pattern Recognition}, pages 14010--14020, 2022.

\bibitem{dsm}
Jinpeng Wang, Yuting Gao, Ke Li, Xinyang Jiang, Xiaowei Guo, Rongrong Ji, and
  Xing Sun.
\newblock Enhancing unsupervised video representation learning by decoupling
  the scene and the motion.
\newblock In {\em The AAAI Conference on Artificial Intelligence (AAAI)}, 2021.

\bibitem{pace_pred}
Jiangliu Wang, Jianbo Jiao, and Yun-Hui Liu.
\newblock Self-supervised video representation learning by pace prediction.
\newblock In {\em The European Conference on Computer Vision (ECCV)}, August
  2020.

\bibitem{semi_tgfixmatch}
Junfei Xiao, Longlong Jing, Lin Zhang, Ju He, Qi She, Zongwei Zhou, Alan
  Yuille, and Yingwei Li.
\newblock Learning from temporal gradient for semi-supervised action
  recognition.
\newblock In {\em Proceedings of the IEEE/CVF Conference on Computer Vision and
  Pattern Recognition}, pages 3252--3262, 2022.

\bibitem{xie2020unsupervised}
Qizhe Xie, Zihang Dai, Eduard Hovy, Thang Luong, and Quoc Le.
\newblock Unsupervised data augmentation for consistency training.
\newblock {\em Advances in Neural Information Processing Systems},
  33:6256--6268, 2020.

\bibitem{semi_mvpl}
Bo Xiong, Haoqi Fan, Kristen Grauman, and Christoph Feichtenhofer.
\newblock Multiview pseudo-labeling for semi-supervised learning from video.
\newblock In {\em Proceedings of the IEEE/CVF International Conference on
  Computer Vision}, pages 7209--7219, 2021.

\bibitem{semi_cmpl}
Yinghao Xu, Fangyun Wei, Xiao Sun, Ceyuan Yang, Yujun Shen, Bo Dai, Bolei Zhou,
  and Stephen Lin.
\newblock Cross-model pseudo-labeling for semi-supervised action recognition.
\newblock In {\em Proceedings of the IEEE/CVF Conference on Computer Vision and
  Pattern Recognition}, pages 2959--2968, 2022.

\bibitem{seco}
Ting Yao, Yiheng Zhang, Zhaofan Qiu, Yingwei Pan, and Tao Mei.
\newblock Seco: Exploring sequence supervision for unsupervised representation
  learning.
\newblock {\em arXiv preprint arXiv:2008.00975}, 2020.

\bibitem{zbontar2021barlow}
Jure Zbontar, Li Jing, Ishan Misra, Yann LeCun, and St{\'e}phane Deny.
\newblock Barlow twins: Self-supervised learning via redundancy reduction.
\newblock In {\em International Conference on Machine Learning}, pages
  12310--12320. PMLR, 2021.

\bibitem{s4l}
Xiaohua Zhai, Avital Oliver, Alexander Kolesnikov, and Lucas Beyer.
\newblock S4l: Self-supervised semi-supervised learning.
\newblock In {\em Proceedings of the IEEE/CVF International Conference on
  Computer Vision}, pages 1476--1485, 2019.

\bibitem{zhang2021flexmatch}
Bowen Zhang, Yidong Wang, Wenxin Hou, Hao Wu, Jindong Wang, Manabu Okumura, and
  Takahiro Shinozaki.
\newblock Flexmatch: Boosting semi-supervised learning with curriculum pseudo
  labeling.
\newblock {\em Advances in Neural Information Processing Systems},
  34:18408--18419, 2021.

\bibitem{zhang2012privacy}
Chenyang Zhang, Yingli Tian, and Elizabeth Capezuti.
\newblock Privacy preserving automatic fall detection for elderly using rgbd
  cameras.
\newblock In {\em International Conference on Computers for Handicapped
  Persons}, pages 625--633. Springer, 2012.

\bibitem{actionreco_vidtr}
Yanyi Zhang, Xinyu Li, Chunhui Liu, Bing Shuai, Yi Zhu, Biagio Brattoli, Hao
  Chen, Ivan Marsic, and Joseph Tighe.
\newblock Vidtr: Video transformer without convolutions.
\newblock In {\em Proceedings of the IEEE/CVF International Conference on
  Computer Vision}, pages 13577--13587, 2021.

\bibitem{hacs}
Hang Zhao, Antonio Torralba, Lorenzo Torresani, and Zhicheng Yan.
\newblock Hacs: Human action clips and segments dataset for recognition and
  temporal localization.
\newblock In {\em Proceedings of the IEEE International Conference on Computer
  Vision}, pages 8668--8678, 2019.

\bibitem{zou2021learning}
Yuliang Zou, Jinwoo Choi, Qitong Wang, and Jia-Bin Huang.
\newblock Learning representational invariances for data-efficient action
  recognition.
\newblock {\em arXiv preprint arXiv:2103.16565}, 2021.

\bibitem{semi_actorcutmix}
Yuliang Zou, Jinwoo Choi, Qitong Wang, and Jia-Bin Huang.
\newblock Learning representational invariances for data-efficient action
  recognition.
\newblock {\em arXiv preprint arXiv:2103.16565}, 2021.

\end{thebibliography}
}
\clearpage

\appendix
\section{Overview}
\begin{itemize}
    \item Section~\ref{sec:impl_semi_supp}: Implementation details about network architectures and training setup. 
    \item Section~\ref{sec:ablation_semi_supp}: Ablation study for our framework.
    \item Section~\ref{sec:method_semi_supp}: Supportive diagrams and explanation for our method.

\end{itemize}

\section{Implementation Details}
\label{sec:impl_semi_supp}
\subsection{Network Architecture}
\subsubsection{Backbone}
For teacher models $f_I$ and $f_D$, we utilize 3D-ResNet50 model from the implementation of \texttt{Slow-R50}~\cite{slowfast} of official PyTorchVideo\footnote{\href{https://github.com/facebookresearch/pytorchvideo}{https://github.com/facebookresearch/pytorchvideo}}. For experiments with 3D-ResNet18, we utilize its official PyTorch implementation \texttt{r3d\_18}\footnote{\href{https://github.com/pytorch/vision/blob/main/torchvision/models/video/resnet.py}{https://github.com/pytorch/vision/blob/main/torchvision/models/video}}. 
\subsubsection{Non-Linear Projection Head}
We use non-linear projection head $g(\cdot)$ during the self-supervised pretraining of temporally-invariant and temporally-distinctive teachers to reduce the dimensions of the representation. We utilize Multi-layer Perceptron (MLP) as a non-linear projection head to project 2048-dimensional model features to 128-dimensional vectors in normalized representation space. The design of MLP is as follows, where \texttt{nn} indicates \texttt{torch.nn} PyTorch package:
\begin{verbatim}
nn.Linear(2048,512, bias = True)
nn.BatchNorm1d(512)
nn.ReLU(inplace=True)
nn.Linear(512, 128, bias = False)
nn.BatchNorm1d(128)
\end{verbatim}

\subsection{Training Details}
For all weight updates, we utilize Adam Optimizer~\cite{adam} with default parameters $\beta_1=0.9$ and $\beta_2=0.999$ with a base learning rate ($\alpha_I$, $\alpha_D$, $\alpha_S$) of 1e-3. For all training, we utilize a linear warmup of 10 epochs. A patience-based learning rate scheduler is also used, which drops the learning rate to its half value on a loss plateau.

\section{Additional Ablations}
\label{sec:ablation_semi_supp}

\subsection{Loss function for teacher supervision}
In order to distill teacher knowledge, we study three different loss functions as $\mathcal{L}_{unsup}$ and report the results in Table~\ref{table:abl_teacherloss}. For these experiments, we use 3D-Resnet50 as the student model on the UCF101 dataset~\cite{ucf101}. We observe that all three losses perform reasonably while $\mathcal{L}_2$ performs the best, which we use as the default loss in our method. 
\begin{table}[h]
\centering
\begingroup
\setlength{\tabcolsep}{6pt}
\begin{tabular}{lccc} 
\arrayrulecolor{black}\hline

\hline

\hline\\[-3mm]
\multirow{2}{*}{\textbf{Unlabeled Supervision}} & \multicolumn{3}{c}{\textbf{UCF101 \% Labels}}  \\
                                                & \textbf{5\%} & \textbf{20\%} & \textbf{50\%}   \\ 
\hline
$\mathcal{L}_2$                                              & 53.48        & 83.15         & 85.02           \\
KL-Divergence                                   & 52.62        & 82.76         & 84.50               \\
JS-Divergence                                             & 50.91            & 82.10             & 83.94               \\
\arrayrulecolor{black}\hline

\hline

\hline\\[-3mm]
\end{tabular}
\endgroup
\caption{Ablation of different Teacher Losses. $\mathcal{L}_2$ distillation loss performs the best, which we use in our default setting.}
\label{table:abl_teacherloss}
\end{table}

\subsection{Student $f_S$ from Scratch}
We perform experiments with student from random initialization and compare them with the prior methods in Table~\ref{table:scratch}.

\begin{table*}[h]
\centering
\begingroup
\setlength{\tabcolsep}{4pt}
\begin{tabular}{ll|cc|cc|c} 

\arrayrulecolor{black}\hline

\hline

\hline\\[-3mm]
                                                                                  & \multirow{2}{*}{Backbone} & \multicolumn{2}{c|}{\textbf{UCF101}}          & \multicolumn{2}{c|}{\textbf{HMDB}}            & \textbf{Kinetics}     \\
                                                                                  &                           & 5\%                   & 20\%                  & 40\%                  & 60\%                  & 10\%                  \\ 
\hline
Prior best methods                                                                        & R3D-50                    & 27.0~\cite{semi_actorcutmix}             & 51.7~\cite{semi_actorcutmix}              & 32.9~\cite{semi_actorcutmix}              & 38.9             & 58.4~\cite{semi_cmpl}             \\
Ours (scratch student)                          & R3D-50                    & 53.1(\textbf{+26.1}) & 83.0 (\textbf{+31.3}) & 52.1 (\textbf{+19.2}) & 54.3 (\textbf{+15.4}) & 60.8 (\textbf{+2.4})  \\ 
\hline
Prior best methods                                                                        & R3D-18                    & 44.8~\cite{semi_tgfixmatch}            & 76.1~\cite{semi_tgfixmatch}             & 46.5~\cite{semi_tgfixmatch}            & 49.7~\cite{semi_tgfixmatch}             & 53.7~\cite{semi_cmpl}             \\
Ours (scratch student)                          & R3D-18                    & 46.7 (\textbf{+1.9})  & 78.2 (\textbf{+2.1})  & 49.1 (\textbf{+2.6})  & 52.9 (\textbf{+3.2})  & 54.4 (\textbf{+0.7})  \\
\arrayrulecolor{black}\hline

\hline

\hline\\[-3mm]
\end{tabular}
\endgroup
\vspace{-2.5mm}
\caption{Experiments with Student trained from Random Initialization. \textbf{(+n)} shows absolute improvement over the prior best work
}
\label{table:scratch}
\end{table*}

\begin{figure*}[h]
    
    \begin{subfigure}{0.49\textwidth}
        \centering
        \includegraphics[height=2.5in]{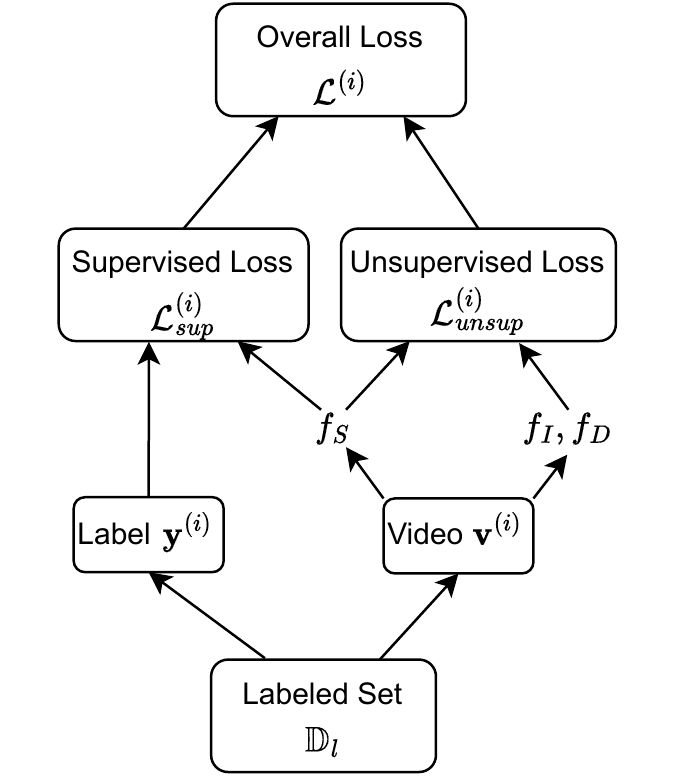}
        \caption{Labeled data}

    \end{subfigure}
    \begin{subfigure}{0.49\textwidth}
        \centering
        \includegraphics[height=2.5in]{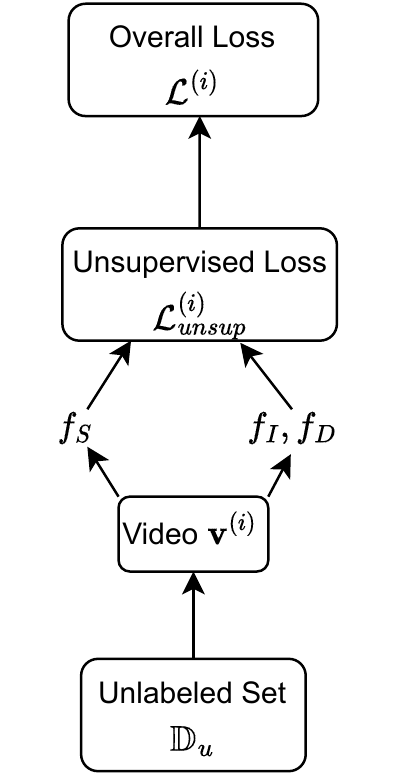}
        \caption{Unlabeled data}

    \end{subfigure}

    \caption{Loss computations in labeled and unlabeled data. (a) In case of Labeled data, the student $f_S$ gets supervision from supervised cross-entropy loss from label $\mathbf{y}^{(i)}$ and unsupervised $\mathcal{L}_2$ loss from teachers. (b) For unlabeled set, the student is only trained with the unsupervised loss from teachers. Details are in Sec~\ref{sec:semisup} of the main paper.}
    \label{fig:lab_unlab}
    \vspace{5mm}
\end{figure*}
\section{Method}
\label{sec:method_semi_supp}
\subsection{Loss for Labeled and Unlabeled set} In Fig.~\ref{fig:lab_unlab}, we show the handling of labeled and unlabeled data in the semi-supervised training of student $f_S$. For labeled data $\mathbb{D}_l$, the student model has two sources of supervision: (1) Labeled supervision $\mathcal{L}^{(i)}_{sup}$ in the form of standard cross entropy loss which is computed from the student's prediction and given class label $\mathbf{y}^{(i)}$ (2) Unlabeled supervision $\mathcal{L}^{(i)}_{unsup}$ in the form of $\mathcal{L}_2$ distillation loss computed from the weighted average of predictions of teachers ($f_D$ and $f_I$). For the unlabeled set $\mathbb{D}_u$, the student model gets supervision only in the form of $\mathcal{L}_2$ distillation loss.

\subsection{Temporally-Distinctive pretraining using unpooled features}
\begin{figure*}[t]
    \centering
    \includegraphics[width=.55\linewidth]{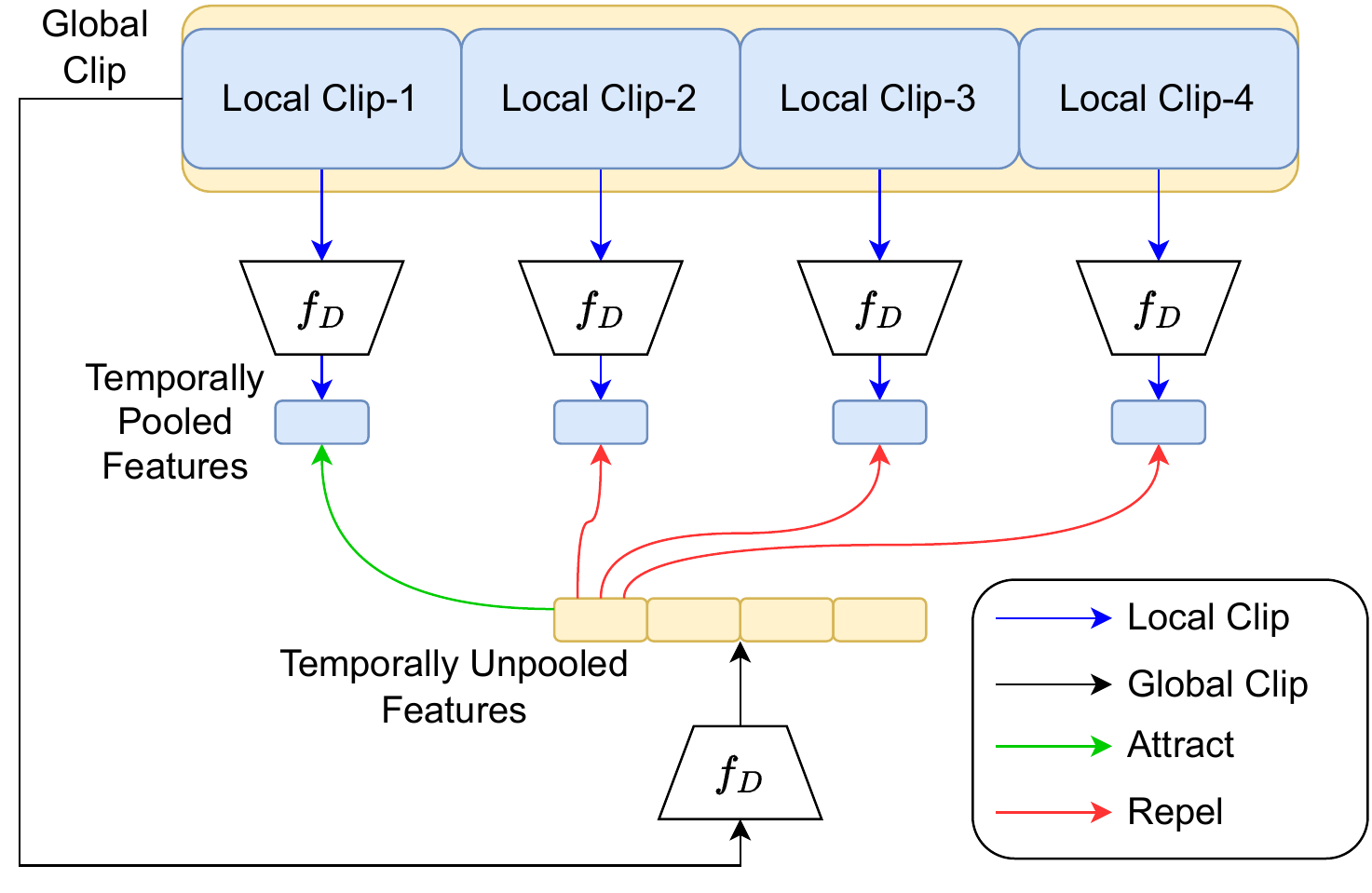}
    \caption{\textbf{Temporally-Distinctive Contrastive Objective for Temporally-unpooled features $\mathcal{L}_{D2}$}: A time-duration of the video can be represented in two different ways: (1) Pooled features of the short(local) clip (2) Unpooled feature slice of the long(global) clip. In this contrastive objective, we maximize the agreement between \textit{temporally-aligned} pooled and unpooled features.}
    
    \label{fig:global_local}
\end{figure*}

Since $\mathcal{L}_{D1}$ deals with temporally-pooled(averaged) features, it promotes temporal-distinctiveness for the \textit{pooled} features. Similar to that, ~\cite{tclr} designs a contrastive objective that promotes temporally-distinctive representation on the \textit{unpooled} features. We call it unpooled temporal-distinctive objective $\mathcal{L}_{D2}$, which is illustrated in Fig.~\ref{fig:global_local}.

\end{document}